\documentclass{article}
\usepackage{graphicx}
\usepackage{microtype}
\usepackage{graphicx}
\usepackage{booktabs} % for professional tables
\usepackage{tabularx}
\usepackage{siunitx}
\usepackage{verbatim}
\usepackage{comment}
\usepackage{xcolor}
\usepackage{subcaption}
\usepackage{makecell}
\usepackage{adjustbox}
\usepackage{wrapfig}  % in your preamble
\usepackage{amsmath}

\usepackage{xcolor}

\usepackage[preprint]{conference}

\usepackage[utf8]{inputenc} % allow utf-8 input
\usepackage[T1]{fontenc}    % use 8-bit T1 fonts
\usepackage{hyperref}       % hyperlinks
\usepackage{url}            % simple URL typesetting
\usepackage{booktabs}       % professional-quality tables
\usepackage{amsfonts}       % blackboard math symbols
\usepackage{nicefrac}       % compact symbols for 1/2, etc.
\usepackage{microtype}      % microtypography
\usepackage{xcolor}         % colors

\title{MetaboNet-Bench: A Multi-modal Benchmark for Glucose Forecasting in Type 1 Diabetes}

\newcommand{\correspondingfootnote}{\thanks{Corresponding authors: \texttt{mpsnyder@stanford.edu}, \texttt{taowang9@stanford.edu }}}
\author {
Nathaniel Jeffries$^{1}$ \\natj@stanford.edu \And
Miriam Wolff$^{2}$ \\miriam@replica.health \And
Sam Royston$^{2}$ \\sam@replica.health \And
Elizabeth Healey$^{3}$ \\elizabeth.healey@childrens.harvard.edu \And
Caleb Mayer$^{1}$ \\mayercl@stanford.edu \And
David Klonoff$^{4}$ \\dklonoff@diabetestechnology.org \And
Michael Snyder$^{1}$ \\mpsnyder@stanford.edu \And
Tao Wang$^{1}$\correspondingfootnote \\taowang9@stanford.edu  \\[2em]
\begin{minipage}{\linewidth}
\raggedright
{$^{1}$Department of Genetics, Stanford University School of Medicine, Stanford, CA, USA} \\[0.3em]
{$^{2}$Replica Health, New York, NY, USA} \\[0.3em]
{$^{3}$Boston Children’s Hospital, Harvard Medical School, Boston, MA, USA} \\[0.3em]
{$^{4}$Diabetes Research Institute, Mills-Peninsula Medical Center, San Mateo, CA, USA}
\end{minipage}
}

\begin{document}

\maketitle

\begin{abstract}

Glucose forecasting algorithms are an important aspect of glycemic control management in type 1 diabetes. So far, the research community has developed numerous algorithms and models for forecasting. However, it is well-recognized that the lack of standardized model performance evaluation benchmarks makes fair comparison difficult and hinders further innovation, and thus benchmark standardization is in urgent need. Furthermore, many published glucose forecasting algorithms are limited to CGM data alone, ignoring other multimodal signals such as insulin dosing and carbohydrate intake. Here, we introduce \textbf{MetaboNet-Bench}, a benchmark for multimodal glucose forecasting for patients with type 1 diabetes that provides an extensible open-source evaluation framework for comparison of glucose forecasting algorithms that leverage glucose, insulin, and carbohydrate data. We then demonstrate its utility by benchmarking several recently published glucose forecasting models and a custom multimodal time-series model, representing different model architectures. The results show that the benefit of adding data modalities is conditioned on the complexity of the model and that incorporating more clinical metrics helps identify meaningful gaps to fill for future research.

\end{abstract}

\section{Introduction}
In recent years, researchers have become increasingly interested in developing blood glucose forecasting algorithms leveraging continuous glucose monitoring (CGM) data. Such algorithms utilize retrospective CGM data, and in some cases, additional information, to predict glucose values at various time horizons. 
Glucose forecasting is an important part of diabetes management for patients with type 1 diabetes (T1D), and many patients with T1D use both CGMs and insulin pumps to manage their diabetes. CGMs produce high-frequency time-series measurements of interstitial glucose concentrations, while insulin pumps record detailed insulin dosing histories that can be leveraged to inform treatment decisions. Glucose forecasting is important not just to improve self-management of diabetes, but also because many automated insulin delivery (AID) systems leverage forecasting models to enhance insulin dosing control. As such, many researchers have recognized the importance of glucose forecasting algorithms \citep{Marx2023-gt, Lara-Abelenda2025-qs, Jaloli2023-gf}. %In addition, glucose forecasting is important for populations beyond T1D patients, for example, prediabetes and type 2 diabetes (T2D) patients, by empowering them to achieve better glycemic control via lifestyle modifications \cite{mike1}. 

While a range of models and model types have been developed in recent years, the dearth of standardized performance benchmarks has made fair model comparison difficult and hindered further innovation. 
Effective standardization requires a full-cycle framework spanning dataset preparation, task definition, and metric selection and reporting to promote reproducibility and reusability. 
Existing efforts address subsets of these needs but remain fragmented, often focusing on data curation, limited evaluations, single-source datasets, or glucose-only forecasting \citep{Prioleau2025-tq, Sergazinov2024}. 
First, the most prominent gap is a lack of a centralized pipeline for evaluating performance results of new models on collections of existing datasets. Without such a platform, it is challenging to understand the benefit of new forecasting models, as performance is greatly influenced by the data used for testing.  A standardized pipeline and open-source implementation of existing models on the same data would improve transparency and rigor in this space. 
Second, a wide range of quantitative performance metrics have been proposed, spanning both model development and clinical management objectives. However, existing benchmarks typically employ a limited subset of these metrics. For example, prior benchmarks often emphasize aggregate performance metrics, leaving model performance with respect to clinically important error characteristics in hypoglycemic versus hyperglycemic ranges understudied \citep{wolff2025blood}. As a result, it is difficult to compare performance in scenarios where forecasting accuracy is critical, such as hypoglycemic ranges.
Third, multimodal data are increasingly more available from individuals, particularly patients with T1D. These data have the potential to improve the performance of glucose-forecasting models. However, the lack of integration of these multimodal data into a standardized data processing pipeline has slowed research in this space. 

To address these limitations, we introduce \textbf{MetaboNet-Bench}, a novel multimodal benchmark for T1D glucose forecasting using insulin, carbohydrate intake and glucose data. MetaboNet-Bench leverages a recently published dataset curation, MetaboNet \citep{wolff2026metabonet}, which includes multimodal data from 13 data sources, facilitating a large-scale evaluation of forecasting algorithms across different datasets.  MetaboNet-Bench is open-source and fully reproducible, with the goal of enabling transparent comparison and accelerating progress in glucose forecasting research. Specifically, this paper makes the following contributions:

\begin{enumerate}
\item We developed an extensible and open-source multimodal benchmark framework, MetaboNet-Bench, that the community can reuse for future model evaluations for glucose forecasting in T1D that extends beyond existing glucose-only benchmarks to incorporate carbohydrates and insulin data. 
\item The utility of MetaboNet-Bench is seen by a systematic evaluation of the performance impact of adding data modalities to state-of-the-art models. Our results show the benefit of adding more data modalities, especially in the presence of common excursion triggers like meals and that models differentially exhibit benefit in adding additional modalities.
\item Our results demonstrate how traditional forecasting metrics like RMSE and MARD can obfuscate performance divergences across glycemic ranges and prandial and correction perturbations, implying work in this space should focus on clinically relevant evaluation metrics.

\end{enumerate}

\section{Related Works}

\subsection{Datasets for Multimodal Glucose Forecasting}

Multimodal glucose forecasting has historically been limited by small, fragmented datasets, which are insufficient for training large machine learning models or performing robust analyses. Early examples, such as the Ohio T1DM dataset \citep{marling2020ohiot1dm}, covered six subjects over six weeks. Subsequent contributions, including BrisT1D \citep{James2025BrisT1DOpenDataset} and DiaTrend \citep{prioleau2023diatrend}, expanded the number of subjects, the duration of recordings, or the feature space, enabling the exploration of new hypotheses. Larger datasets focusing solely on CGM are excluded here because they lack insulin or carbohydrate data, which are critical drivers of glycemic variability in type 1 diabetes. A real-world multimodal dataset (2,217 participants, with roughly 28 days of data per participant), including CGM, wearable (heart rate), and user-logged nutrition, activity, and weight, was collected in the Season of Me program and used to train an LSTM-based CGM prediction model with multimodal inputs \citep{Zahedani2023-sw}. However, the dataset is not publicly available and is not standardized for community benchmarking.

The T1DEXI dataset \citep{riddell2023t1dexi} provides a rich collection of exercise-related data in T1D, suitable for detailed analysis. However, T1DEXI, like many other T1D management datasets, requires an onerous Data Use Agreement and a multi-step process involving human review before temporary data access is provided, limiting accessibility and community engagement for algorithm development. In contrast, our benchmark is the first study to leverage the MetaboNet dataset \citep{wolff2026metabonet}, which consolidates 13 public datasets containing CGM, insulin, carbohydrate intake, additional physiological signals, and demographics.  MetaboNet intends to incorporate new training data with each new release, while keeping the test set static. The public version is easily downloaded from \url{https://metabo-net.org} and includes 1,895 subjects and 1,464 subject-years, enabling frictionless large-scale, standardized evaluation across diverse subpopulations.

\subsection{Forecasting Models}

A variety of blood glucose prediction models have been developed in recent years, ranging from classical time-series approaches to deep learning models incorporating multimodal inputs. For example, CGM-LSM \citep{Luo2025-je} is a transformer-based large sensor model pretrained on over 15.96 million CGM recordings. On the  Ohio T1DM dataset, it achieves approximately 15.9 mg/dL RMSE for 1-hour glucose forecasting. Gluformer \citep{Sergazinov2022-oe} is a transformer-based infinite mixture model with uncertainty quantification that achieves a 15.4 mg/dL RMSE for 1 hour glucose forecasting on the DeepMO Dataset \citep{fox2018deep}. Other approaches have incorporated multimodal inputs, including insulin, carbohydrate intake, and physical activity, to capture additional sources of glycemic variability \citep{yang2020multiscale}. Recent work has also explored general-purpose time series foundation models, such as UniTS \citep{Gao2024-bo}, demonstrating the potential of large, pretrained models to generalize across diverse temporal prediction tasks.

\subsection{Existing Benchmarks and Evaluation Frameworks}

There has been a recognized need for standardizing datasets and benchmarks in diabetes research \citep{Wolff2025-tr}. GlucoBench \citep{Sergazinov2024} introduced a framework for standardized prediction tasks and systematic comparisons between blood glucose prediction methods, but it focuses solely on CGM data. \cite{xie2020benchmarking} benchmarked several models on the Ohio T1DM dataset, comparing various feature combinations. They note that the OhioT1DM dataset may not ensure sufficient individual input excitation, which can limit model training. GluPredKit \citep{Wolff2024jossglupredkit} is a Python package that supports data processing and evaluation pipelines for blood glucose prediction, but it focuses more on end-to-end training and evaluation rather than direct model-to-model comparison. Other approaches have employed compartment models and digital twins \citep{cappon2023replaybg, cappon2023individualized}. Existing benchmarks primarily rely on aggregate accuracy metrics, such as RMSE, which provide limited insight into clinical applicability \citep{wolff2025blood}. Consequently, most evaluations ignore clinically and scientifically relevant dimensions, such as performance across glycemic regions, demographics, or subpopulations.

Our benchmark addresses these limitations by jointly evaluating multimodal glucose forecasting models at scale while enabling clinically meaningful subgroup analyses. Leveraging the consolidated MetaboNet dataset enables comparisons across models, datasets, and metrics, including stratified analyses previously infeasible due to data size or fragmentation.

\section{Methods}

\begin{wrapfigure}{htb}{0.45\columnwidth}
\centering
\includegraphics[width=0.43\columnwidth]{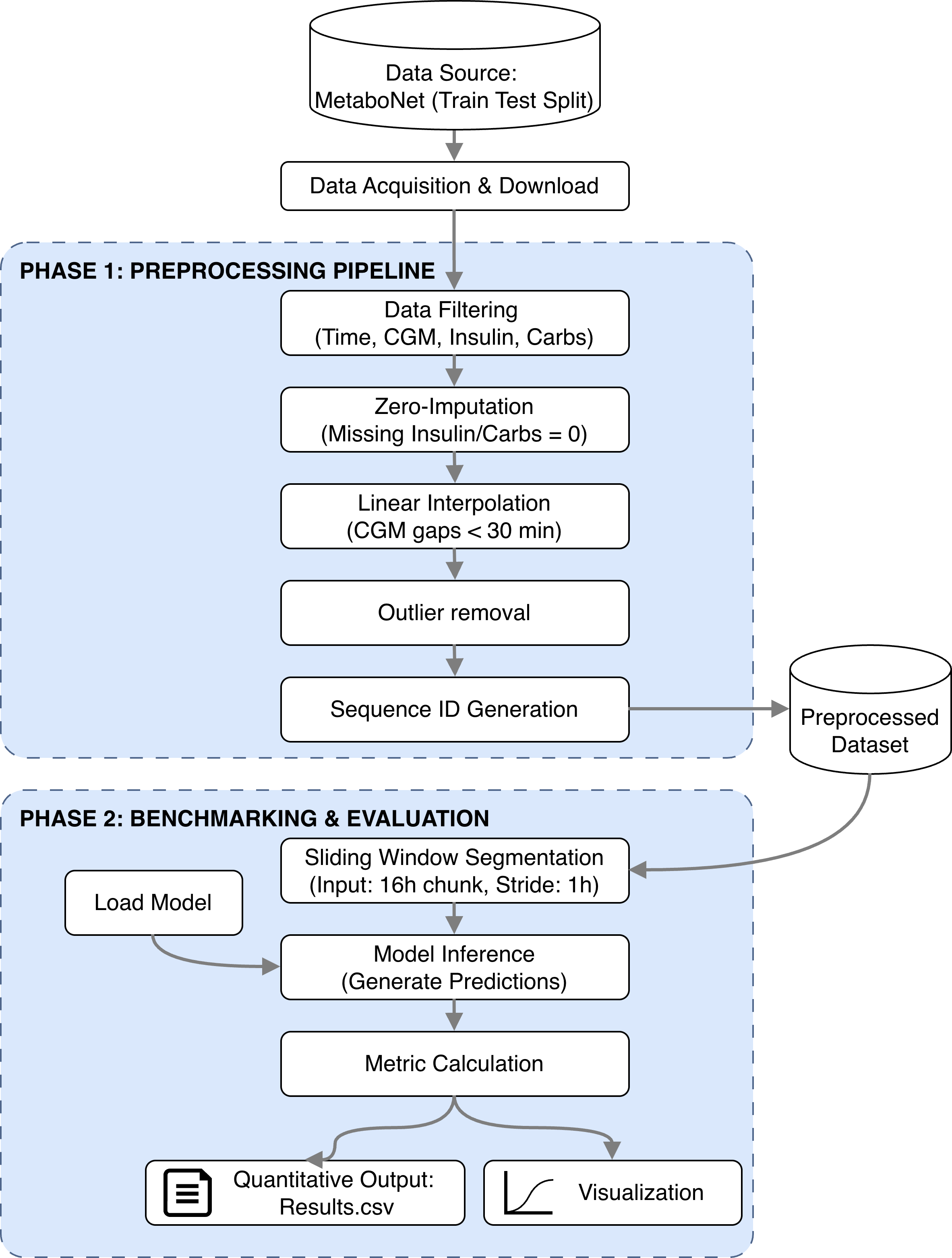}
\caption{Workflow of MetaboNet-Bench, showing data retrieval, preprocessing, sliding-window segmentation, model inference, and evaluation for glucose forecasting.}
\label{fig:workflow}
\end{wrapfigure}

Figure \ref{fig:workflow} illustrates the MetaboNet-Bench workflow. Data are retrieved and preprocessed by filtering features, imputing zeros for missing insulin and carbohydrate values, and removing outliers. The data are then segmented using a sliding window before model inference. Finally, models are evaluated on the glucose forecasting task using quantitative metrics and visualizations of clinical accuracy.

\subsection{Datasets}

This study is the first to use the public subset of the MetaboNet dataset \citep{wolff2026metabonet}, which consolidates 13 datasets listed in Table \ref{tab:datasets}. The inclusion criteria for this dataset are overlapping CGM and insulin data generated by T1D patients. Additional features, such as physical activity data and patient demographics, were included when available. Furthermore, we excluded the private subset of MetaboNet containing restricted Data Use Agreement–Governed (DUA-Governed) datasets to facilitate easy access and promote community engagement.  MetaboNet is unrestricted for research use via a license available on \url{https://metabo-net.org}. A complete description of the features and instructions for data download are available on the MetaboNet website \citep{metabonet_data_dictionary}.

GlucoBench \citep{Sergazinov2024} conducted their study on five datasets, comprising a total of 461 subjects. As GlucoBench focused exclusively on glucose data, none of their datasets overlap with those used in this study due to MetaboNet’s inclusion criteria. Glucose-ML \citep{Prioleau2025-tq} incorporates datasets totaling 2,559 subjects and 38M samples, but only the Shanghai T1DM dataset \citep{zhao2022_shanghaiT1DM_T2DM} meets the inclusion criteria of MetaboNet. Some of the datasets used in Glucose-ML are part of MetaboNet’s private subset, which was intentionally excluded from this study because it is not publicly accessible. In total, the public MetaboNet dataset we used for this benchmark includes over 153M samples \citep{wolff2026metabonet}.

\subsection{Pre-processing}

\textbf{Train/Test Split and Task Definitions:}
The MetaboNet dataset can be downloaded as separate training and test sets from \url{https://metabo-net.org}, with each individual dataset component deliberately represented in both splits. The test set comprises approximately 15\% of the total dataset. First, the split is performed by patient, so that roughly 10\% of the subjects in each individual dataset are assigned exclusively to the test set; this split is hereafter referred to as the \textit{novel patients split} (\textbf{Task 1}), which is used to evaluate how well a model can perform on completely new patients. Second, 10\% of the subjects from the remaining training set are further divided, with the time-sorted second half of their data assigned to the test set. These subjects will be referred to as the \textit{known patients split} (\textbf{Task 2}), which is used to evaluate how well a model can incorporate prior data from a patient to improve future performance accuracy. None of the models here were designed to explicitly take advantage of this construction. The main results report the full test set, combining both known patient splits and novel patient splits.

\textbf{Dataset Filtering:}
Some datasets in MetaboNet are missing carbohydrate values between meals, as shown in Table \ref{tab:datasets}, and consequently were excluded from this study. We filled up to 12 hours of missing carbohydrate values with 0 following a valid carbohydrate entry to avoid filtering valid rows. We filled up to 1 hour of missing insulin values with 0 following a valid insulin entry. Next, we filtered all rows missing timestamps, carbohydrate data, and insulin. The datasets contain patients using both insulin pumps and multiple daily injections (MDI). We filtered out all patients on MDI since long-acting insulin causes a response lasting beyond the 15-hour context used within MetaboNet-Bench. Overall, filtering reduced the number of per-patient time slices in the dataset from 22.1M to 17.3M.

\textbf{Interpolation and Sequence Segmentation:}
Some datasets use CGMs with a longer than 5 minute reading interval, or contain short sequences of missing CGM values. We performed linear interpolation up to 30 minutes between valid CGM readings to address discrepancies in CGM interval and to fill in short CGM signal losses. The benchmark evaluates models on 15 hours of context with prediction horizons up to 1 hour. We identified remaining gaps in data and created a sequence ID where each contiguous sequence has a unique ID and filtered out sequences less than 16 hours to simplify benchmark logic.

\subsection{Models}

We chose a diverse range of models, from simple statistical models, general purpose forecasting, glucose-only prediction models and multimodal glucose, insulin and carbohydrate models across a range of model architectures.  Appendix Table \ref{tab:data} differentiates the input data modalities of each model. Appendix Table \ref{tab:complexity} includes a comparison of model size and complexity.  Below, we include a basic description of each model. The complete implementation descriptions and hyperparameter selection and tuning procedure, performed using Optuna \citep{Akiba2019-pj}, is included Appendix \ref{app:methods}.

\textbf{Baseline Models:}
To establish a baseline, we evaluated a zero order hold (ZOH) model and a simple linear extrapolation (LE). These models were designed using glucose data alone. The ZOH model predicts the last observed glucose value for every time horizon. The LE model computes the slope from the past 15 minutes and extrapolates forward from the latest glucose value.

\textbf{Light Gradient-Boosting Machine:}
Light Gradient-Boosting Machine (LightGBM) is an open-source gradient boosting framework. This model has tabular inputs with time-lagged features with up to two hours of history. 

\textbf{Ridge:}
Ridge is a linear regressor with L2-regularization. This model has the same inputs as LightGBM, including time-lagged features up to two hours of history. 

\textbf{Gluformer:}
We include a domain-specific forecasting model, Gluformer \citep{Sergazinov2022-oe}, which has been previously benchmarked \citep{Sergazinov2024}. Gluformer is a foundational blood glucose prediction model 
\citep{Sergazinov2022-oe}. 

\textbf{UniTS:}
UniTS is a foundational time series model with forecasting capabilities \citep{Gao2024-bo}. UniTS is pre-trained on a variety of data domains, including human activity sensors, healthcare, engineering, and finance. 

\textbf{Custom LSTM:}
We included a two-layer LSTM as a baseline. This model was trained on the MetaboNet training split using an identical training loop to Gluformer but with MSE loss. The LSTM model has a hidden size of 128 with two LSTM layers and a sequence length of 180 and one head to predict the 12 horizons.

\textbf{GluForecast:}
To leverage the multimodality of the benchmark, we included a novel transformer model, designed to leverage time, blood glucose, insulin, and carbohydrates. The model is a decoder-only sequence-to-sequence model with 12 heads - one for each prediction horizon up to an hour. We use a causal architecture and calculate MSE loss using the input shifted by each head's prediction horizon. Loss is calculated against model predictions with at least 3 hours (36 values) of context. This served as a custom transformer baseline for comparison.

\subsection{Evaluation}

\textbf{Prediction Tasks:} The benchmark reports model accuracy for each 5-minute interval up to 60 minutes ahead. We selected 30 minutes as the primary horizon due to its clinical relevance in the control algorithm for artificial insulin delivery (AID) systems. This prediction horizon is also widely used across literature \citep{Calzavara2026-jl}.

\textbf{Evaluation Metrics:} The benchmark reports accuracy in terms of conventional metrics, including root mean squared error (RMSE) and mean absolute relative difference (MARD). These metrics have conventionally been used in glucose forecasting \citep{Calzavara2026-jl}, over other metrics such as Pearson correlation, as they better capture the magnitude of error. We additionally include clinical metrics to capture performance characteristics relevant to type 1 diabetes.  These include the \textit{Diabetes Technology Society (DTS)} error grid \citep{dts-error-grid}, which differentially penalizes errors based on both the reference value and the predicted value, and on \textit{system perturbations} (including \textit{postprandial} and \textit{insulin corrections}). Appendix \ref{app:metrics} details the metric definitions and includes justification for their choice. 

\textbf{Ablation Study:}
We conducted ablation studies on the following models: LSTM, GluForecast, UniTS, LightGBM, and Ridge. For each model, we included the same preprocessed CGM and timestamp data but varied the inclusion of insulin and carbohydrate data into the feature set, leading to 4 versions for each of 5 models including CGM only, CGM and carbs, CGM and insulin, and CGM, insulin and carbs. Hyperparameters were tuned for each model, detailed in the Appendix, and the best performing model was shown.

\subsection{Extensibility and Reproducibility}

MetaboNet-Bench is entirely open-source and code is available on Github: \url{https://anonymous.4open.science/r/MetaboNet-Bench-4FAD/README.md}. All models evaluated are available on Huggingface Hub or within the Github repo where they can be freely evaluated and run. The benchmark is designed to make incorporating new models extremely easy. A new Huggingface hub model can be added by adding one model implementation file and adding the model to the list of supported models. Implementation files for existing models are typically 25 lines of code or less.

\section{Results}

This section presents the results of MetaboNet-Bench. We first report overall aggregate metrics on the full dataset, as well as for the novel patient split  and known patient split. Next, we assess clinical performance using the DTS error grid. Finally, we provide a detailed analysis across different glycemic ranges and subpopulations.

\subsection{Performance on Standard Metrics}

Figure 2 shows the results from the entire test set and the DTS error grid visual for the LE and GluForecast models. Table \ref{tab:overall-metrics} shows the full results. For short-term prediction horizons of up to 15 minutes, the LE demonstrated comparable performance and, in some cases, outperformed several more complex models; however, its relative efficacy declined for longer prediction horizons. The baseline ZOH model was consistently surpassed by all other models, with the exception of the 5-minute horizon. GluForecast showed superior performance across RMSE, MARD, and percent time in DTS zone A. 

\begin{figure}[htbp]
\centering
\begin{small}

\includegraphics[width=0.3\linewidth]{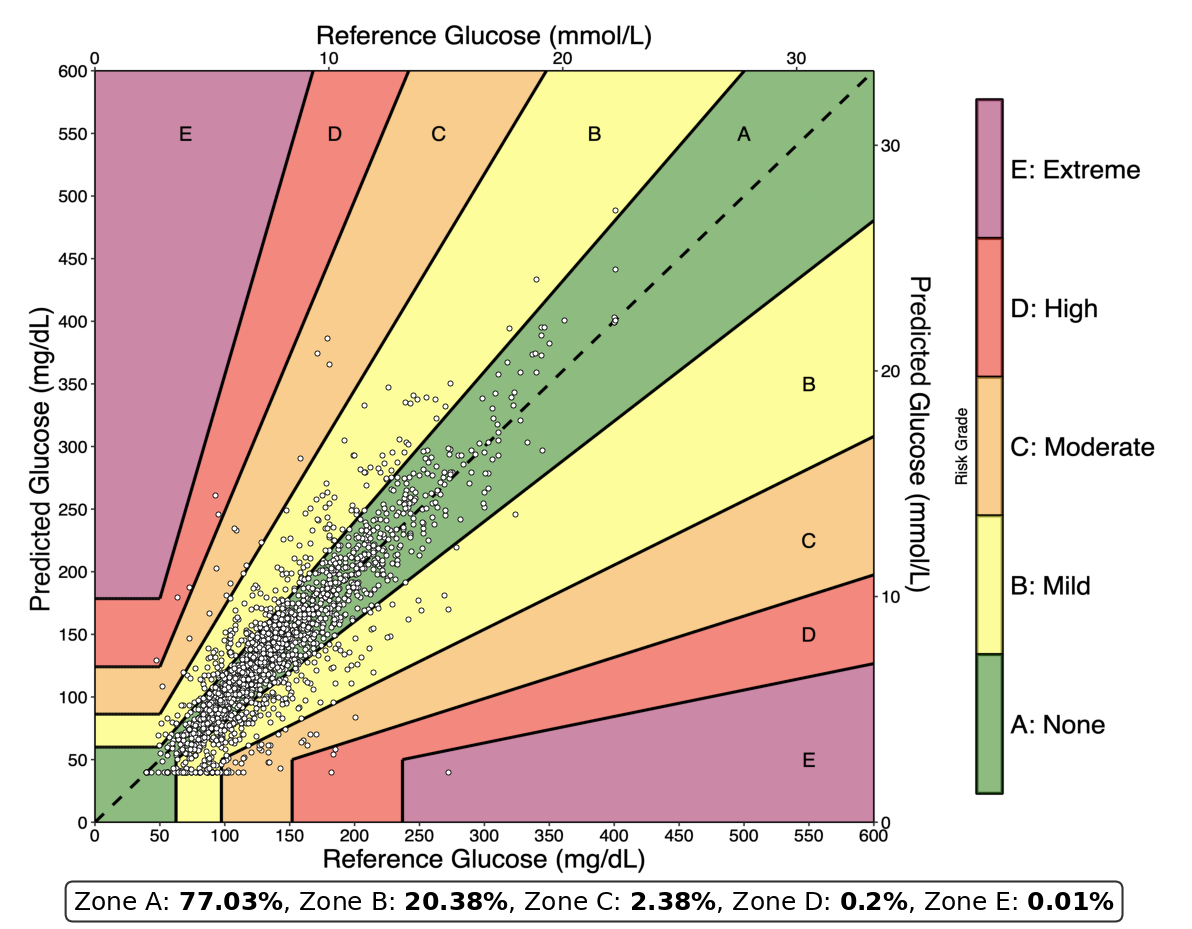}
\hfill
\includegraphics[width=0.3\linewidth]{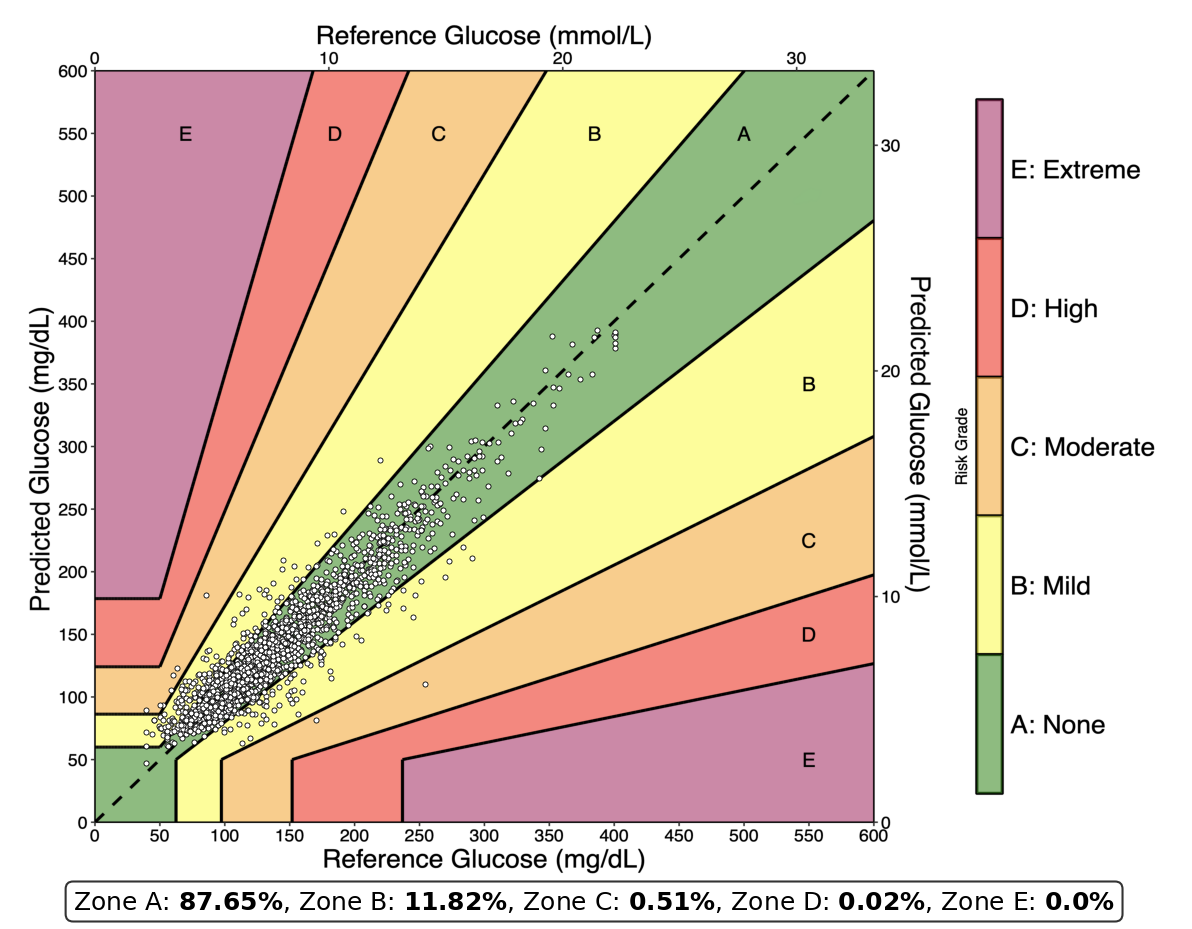}
\hfill
\includegraphics[width=0.3 \linewidth]{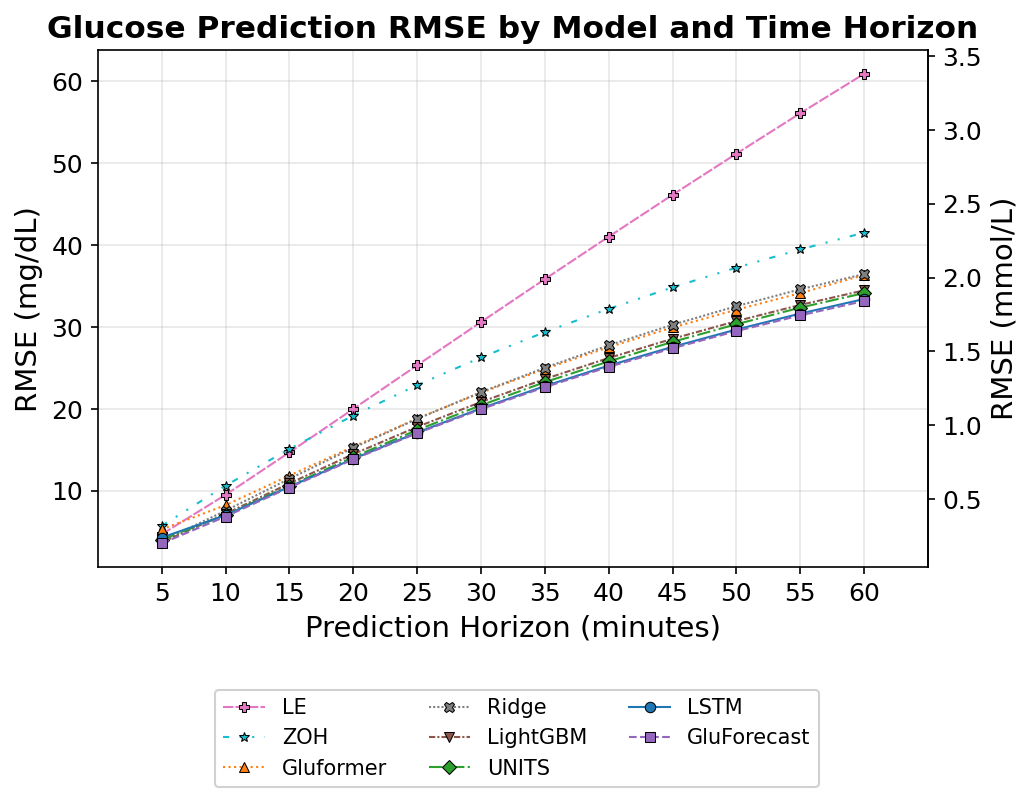}

\caption{
(Left) Linear extrapolation DTS error grid at 30-minute PH.
(Middle) GluForecast DTS error grid at 30-minute PH.
(Right) RMSE across all models and prediction horizons in MetaboNet-Bench. %Predictions outside the green zone are considered clinically inaccurate
For DTS grids, predictions outside the green zone are considered clinically inaccurate, with colored zones indicating clinical risk.}
\label{fig:metrics-overview}
\end{small}
\end{figure}

Figure \ref{fig:task-1-2} presents the results for the novel patient and known patient splits, respectively. The results were largely consistent with the overall performance, with no changes in the ranking of models across metrics or prediction horizons. Slightly higher accuracy was observed in the known patient splits, but this might be an artifact of glycemic variability within each group, since the naïve ZOH and LE models also showed the same pattern. A granular overview is presented in Tables \ref{tab:person-independent-split} and \ref{tab:fine-tuning-split}.

We also evaluated the signed error, by illustrating a ridge plot to assess forecasting biases and visualize errors across models. As shown in Figure \ref{fig:ridge-signed-error}, the signed error distribution was generally centered around zero, as expected, although variability increases for longer horizons. Notably, GluForecast exhibited both narrower error bounds and superior performance across prediction horizons while the LSTM model consistently overpredicted blood glucose values across all horizons, indicating limitations in its ability to generalize across diverse subject distributions.
\subsection{Clinical Evaluation of Predictions}

\textbf{DTS Error Grid Analysis:} We employ the DTS error grid to assess model performance in a clinical context. Figure \ref{fig:metrics-overview} presents the DTS error grids for Linear Extrapolation and GluForecast, respectively. Readings within the green A zone are considered clinically accurate, while darker colors indicate higher clinical risk. Table \ref{tab:dts-error-grid-metrics} summarizes the distribution of predictions across each DTS zone for all models at a 30-minute prediction horizon. These results further demonstrate that GluForecast achieved the best overall performance, both in conventional metrics and in clinically relevant terms.

\begin{table*}[t]

\centering
\caption{Model performance with 95\% bootstrap confidence intervals (1{,}000 resamples) and DTS error grid zones for a 30-minute prediction horizon.}.
\label{tab:dts-error-grid-metrics}
\footnotesize

\resizebox{\textwidth}{!}{%
\begin{tabular}{lccccc}
\toprule
Model & RMSE (mg/dL) & MARD (\%) & DTS A (\%) & DTS B (\%) & DTS C--E (\%) \\
\midrule
GluForecast & \textbf{19.95 [19.91, 20.00]} & \textbf{10.76 [10.74, 10.79]} & \textbf{87.65 [87.59, 87.71]} & 11.82 [11.76, 11.88] & 0.53 [0.52, 0.54] \\
Gluformer   & 21.97 [21.92, 22.02] & 11.69 [11.66, 11.72] & 85.69 [85.62, 85.75] & 13.67 [13.61, 13.74] & 0.64 [0.63, 0.65] \\
LightGBM    & 20.84 [20.79, 20.89] & 11.27 [11.25, 11.30] & 86.55 [86.49, 86.61] & 12.84 [12.78, 12.90] & 0.61 [0.59, 0.62] \\
LE          & 30.60 [30.53, 30.68] & 15.54 [15.51, 15.57] & 77.03 [76.95, 77.11] & 20.38 [20.31, 20.45] & 2.59 [2.56, 2.62] \\
LSTM        & 20.06 [20.01, 20.11] & 10.89 [10.87, 10.91] & 87.44 [87.37, 87.50] & 12.03 [11.98, 12.10] & 0.53 [0.51, 0.54] \\
Ridge       & 22.03 [21.98, 22.08] & 11.94 [11.91, 11.96] & 85.34 [85.28, 85.41] & 13.84 [13.78, 13.90] & 0.82 [0.80, 0.83] \\
UniTS       & 20.44 [20.39, 20.49] & 10.93 [10.91, 10.95] & 87.25 [87.19, 87.31] & 12.18 [12.13, 12.24] & 0.57 [0.56, 0.58] \\
ZOH         & 26.28 [26.22, 26.34] & 13.80 [13.77, 13.82] & 80.77 [80.69, 80.84] & 17.86 [17.79, 17.93] & 1.37 [1.35, 1.39] \\
\bottomrule
\end{tabular}%
}
\end{table*}

\begin{figure*}[t]
\centering
\begin{minipage}[t]{0.29\textwidth}
\centering
\includegraphics[width=\linewidth]{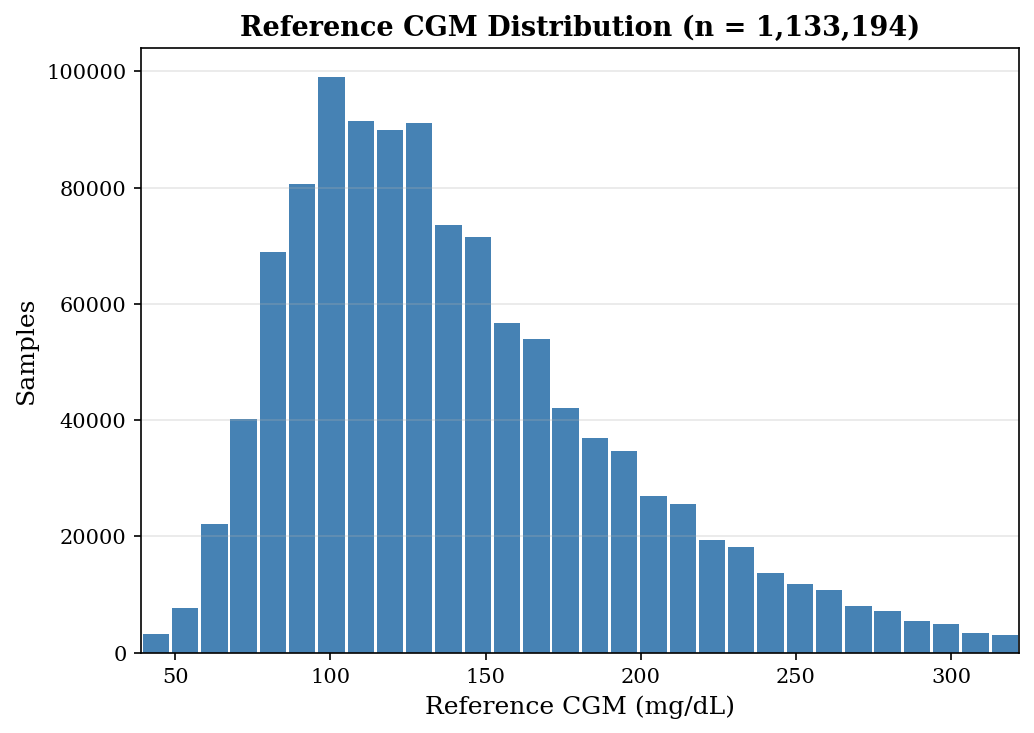}
\subcaption{Distribution of samples across glycemic regions.}
\label{fig:cgm-regions-hist-a}
\end{minipage}
\hfill
\begin{minipage}[t]{0.30\textwidth}
\centering
\includegraphics[width=\linewidth]{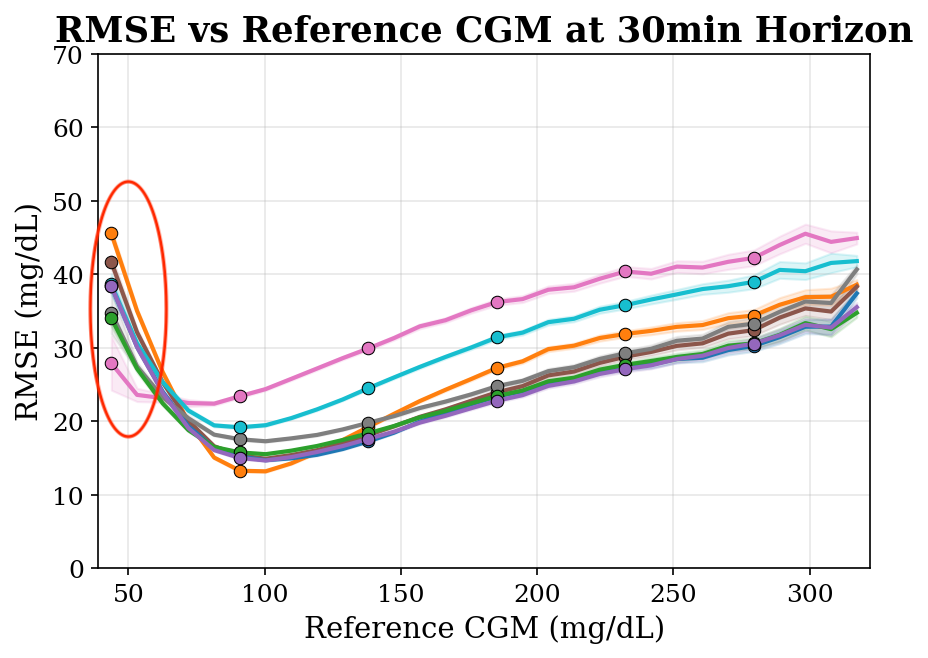}
\subcaption{RMSE by target region for 30-minute prediction horizon.}
\label{fig:cgm-regions-hist-c}
\end{minipage}
\hfill
\begin{minipage}[t]{0.39\textwidth}
\centering
\includegraphics[width=\linewidth]{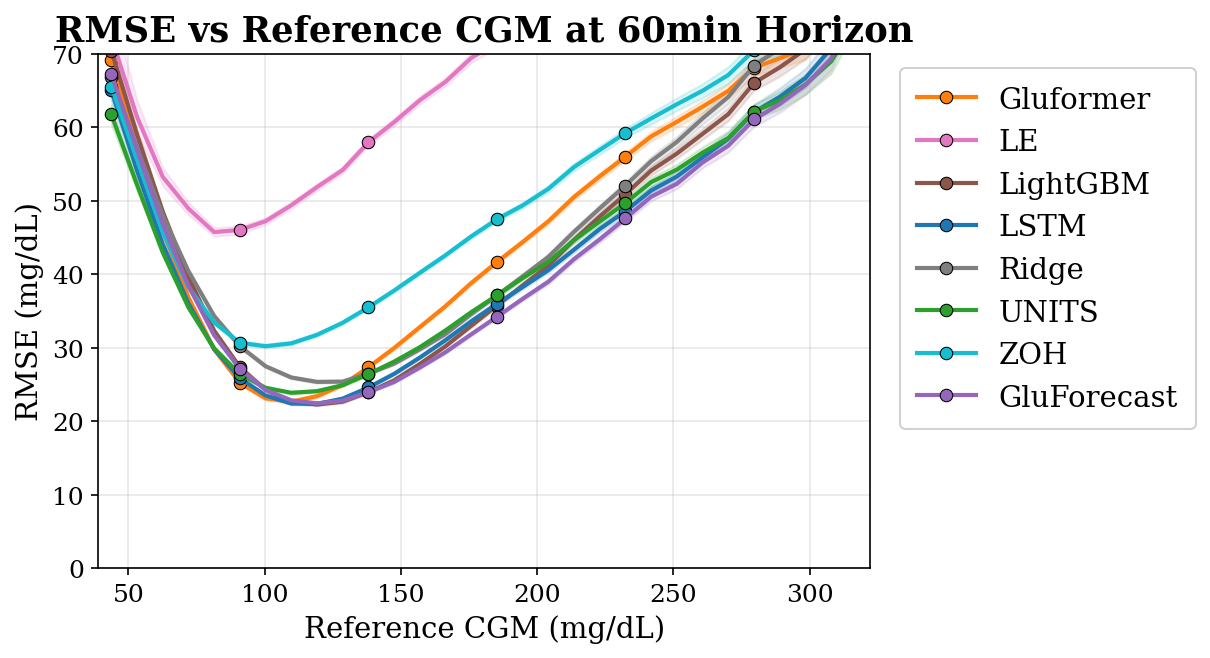}
\subcaption{RMSE by target region for 60-minute prediction horizon.}
\label{fig:cgm-regions-hist-b}
\end{minipage}
\caption{Distribution of samples across glycemic regions (left) and corresponding RMSE results for 30-minute (middle) and 60-minute (right) prediction horizons. RMSE values are computed after filtering predictions by the glycemic regions indicated on the x-axis. The red circle highlights the hypoglycemic region where the LE model outperforms the otherwise consistently superior GluForecast model for 30-minute predictions.}
\label{fig:cgm-regions-hist}
\end{figure*}
\textbf{Evaluation Across Glycemic Ranges:} Data from T1D management are inherently imbalanced, with clinically critical hypoglycemic and hyperglycemic samples underrepresented compared to in-range samples \citep{wolff2025blood}. This reflects the clinical goal of maintaining a percentage of time in range (70–180 mg/dL) above 70\%. Such an imbalance poses a challenge for data-driven models, which may struggle to accurately predict values in the underrepresented glycemic regions. Figure \ref{fig:cgm-regions-hist} illustrates both the distribution of glucose values across glycemic regions and the corresponding RMSE for each model. For prediction horizons up to 30 minutes, the LE was the best performer in the lowest glycemic regions, indicating that data-driven models may not generalize well in these ranges and that simpler models can already achieve strong performance for short horizons. In contrast, longer prediction horizons represent a more challenging task, where more complex models demonstrate advantages. Overall, RMSE was higher in hypoglycemic and hyperglycemic regions, reflecting the increased variability and difficulty of predicting extreme glucose values. More granular results from this analysis are presented in Figure \ref{fig:heatmap-cgm-region}.

\textbf{Evaluation Across Subpopulations:} We conducted an evaluation across multiple subpopulations, which is made possible for the first time by the breadth and diversity of the MetaboNet dataset. Across all examined subpopulations, GluForecast consistently achieved the best predictive performance, with no unexpected reordering of model rankings. Figure \ref{fig:subpopulations_demographics} summarizes performance stratified by age, weight, height, and gender. While overall trends were stable across models, predictive accuracy declined markedly for individuals younger than 20 years across all methods. 

\subsection{Ablation Study}

\begin{figure}
\centering

\includegraphics[width=.8\linewidth]{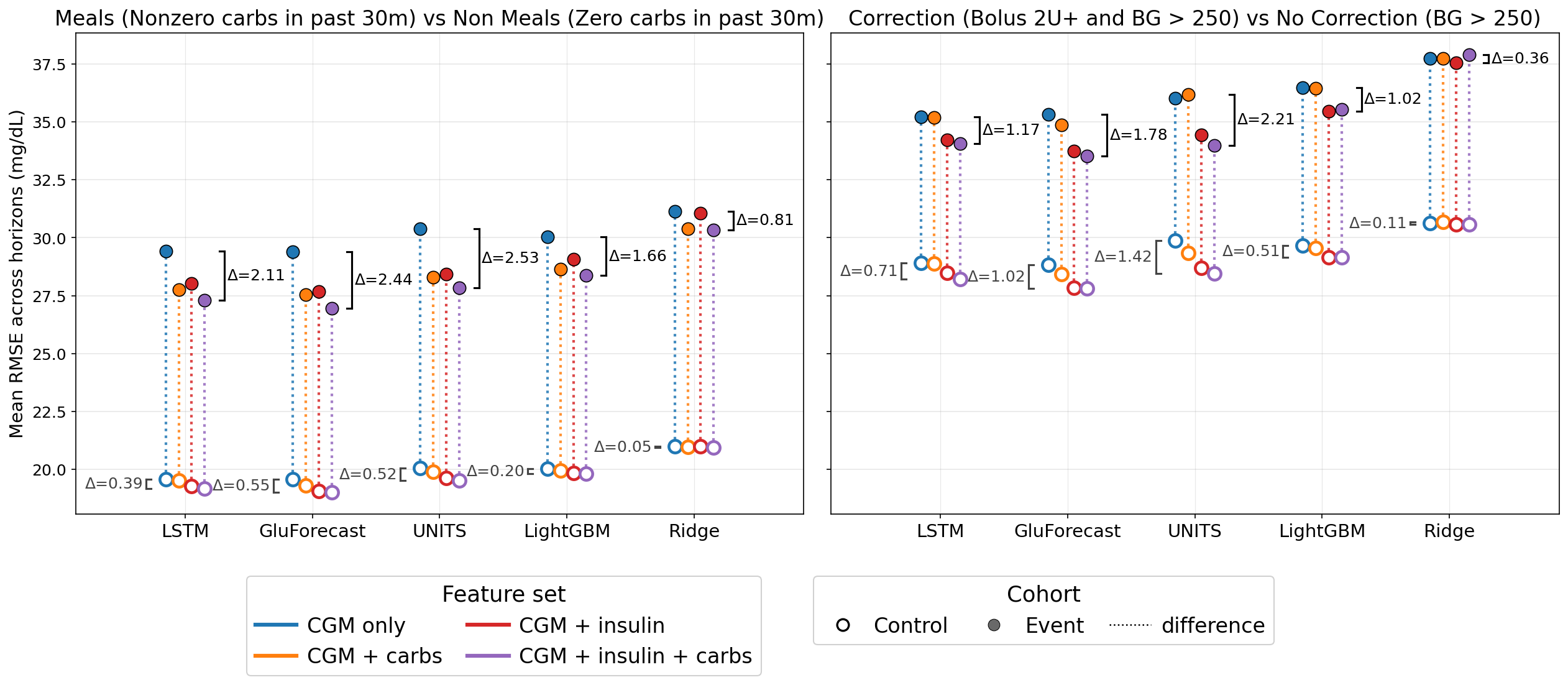}
\caption{Comparison of performance of the ablated models in the presence of common blood glucose perturbations averaged at all horizons. The performance of models is contrasted against a control in the setting of postprandial prediction (left) and hyperglycemic correction boluses (right). The delta shows the difference between the best and worst performing model for the perturbation and the control for each model. }
\label{fig:correction_meal_diff}
\end{figure}

\begin{wrapfigure}{r}{0.5\textwidth}
    \centering
    \includegraphics[width=0.48\textwidth]{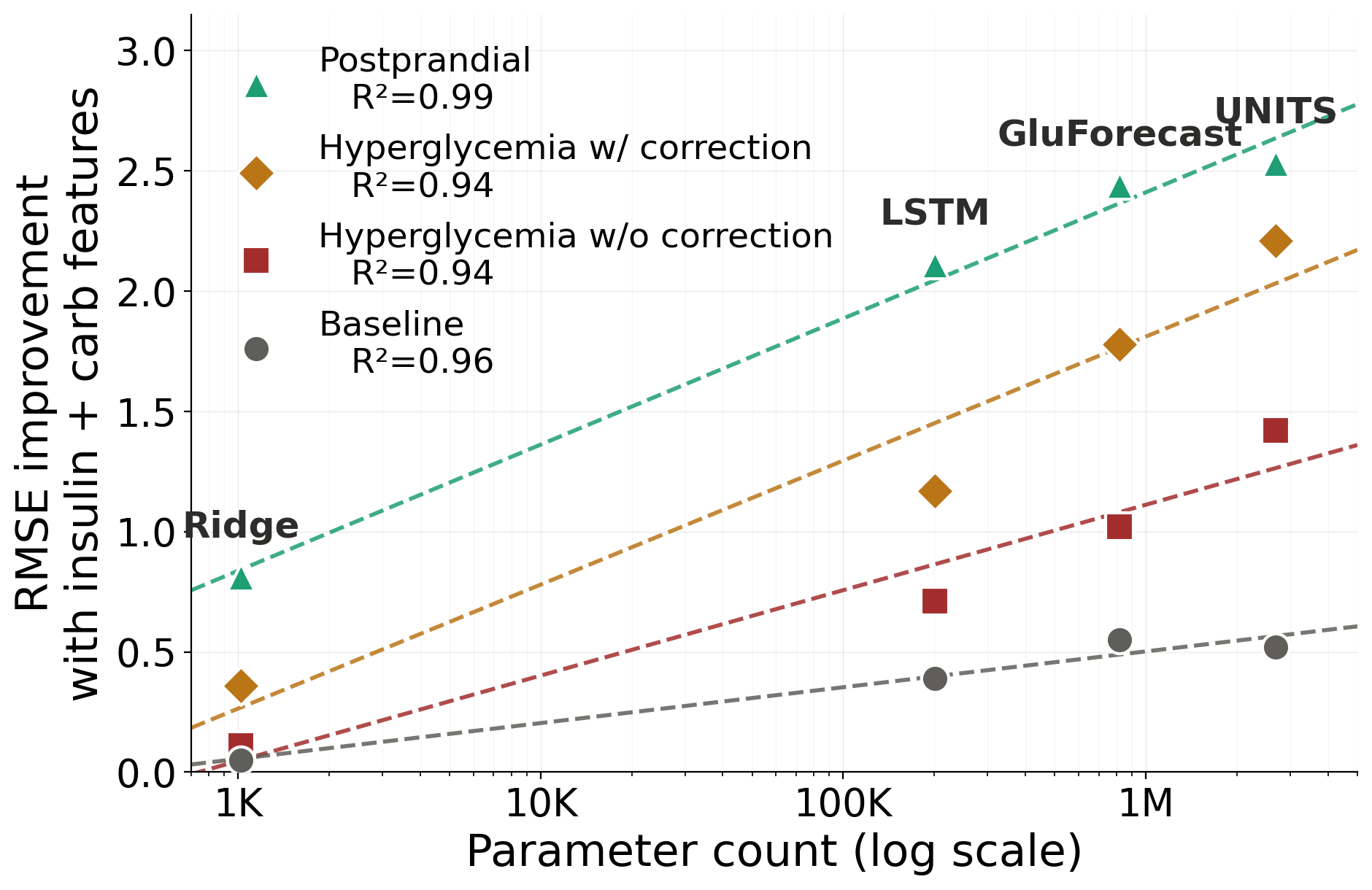}
    \caption{Relation between parameter count and RMSE improvement due to introduction of Insulin and Carbohydrate data  ($\Delta$ in Figure \ref{fig:correction_meal_diff}). Parameter counts are from Table \ref{tab:complexity} in the appendix.  LightGBM was not included in this chart since it is non-parametric and escapes any natural comparison by parameter count.}
    \label{fig:scaling_laws}
\end{wrapfigure}

Table \ref{tab:ablation-study-results} shows the performance results of the ablation study. Overall, multimodal models achieved meaningfully better performance with access to both insulin and carb data.  As shown in Figure \ref{fig:correction_meal_diff} and supplementary figures \ref{fig:mealtime} and \ref{fig:correction}, differences in performance were more apparent postprandially, and to a lesser extent, following correction doses to treat high blood glucose. Intuitively, in Figure \ref{fig:correction_meal_diff} we see that following a meal, carbohydrates appear to be a slightly more useful signal than insulin, while for correction boluses we can see that insulin is much more important.  In general, insulin appears to be a more helpful feature for most models than carbohydrates; this may be partially due to the higher fidelity data from insulin delivery systems which is reported on a 5 minute basis, while carb logs are reported by users a few times a day at best. 

The ablation study also suggests that parameter count may play a role in a model's ability to leverage insulin and carbohydrate data effectively. Though not the top performer in absolute terms, the model with the largest number of trainable parameters (UniTS, see Table \ref{tab:complexity}) also improves the most relative to CGM-only, while on the other hand Ridge Regression, a shallow model, receives hardly any benefits from access to Insulin and Carbohydrate data.

As demonstrated in Figure \ref{fig:scaling_laws}, these observations suggest that in the context of a large scale dataset like MetaboNet, larger models can make bigger gains via multimodality than smaller ones.

\begin{table*}[hbt]
\centering

\caption{Results of ablation study with 95\% bootstrap confidence intervals (1{,}000 resamples) with respect to input features at a 30-minute prediction horizon.}
\label{tab:ablation-study-results}
\resizebox{\textwidth}{!}{%
\begin{tabular}{lcccc}
\toprule
\textbf{Model} & \textbf{cgm} & \textbf{cgm + carb} & \textbf{cgm + insulin} & \textbf{cgm + insulin + carb} \\

\midrule
\multicolumn{5}{c}{\textbf{RMSE} (mg/dL)} \\
\midrule
GluForecast & 20.66 [20.62, 20.71] & 20.30 [20.26, 20.35] & 20.06 [20.02, 20.11] & \textbf{19.95 [19.91, 20.00]} \\
LightGBM    & 21.17 [21.12, 21.22] & 20.97 [20.93, 21.02] & 20.93 [20.88, 20.98] & \textbf{20.84 [20.79, 20.89]} \\
LSTM        & 20.59 [20.54, 20.63] & 20.46 [20.42, 20.51] & 20.16 [20.11, 20.20] & \textbf{20.06 [20.01, 20.11]} \\
Ridge       & 22.16 [22.10, 22.21] & 22.05 [22.00, 22.10] & 22.14 [22.08, 22.19] & \textbf{22.03 [21.98, 22.08]} \\
UniTS       & 21.10 [21.04, 21.15] & 20.81 [20.76, 20.86] & 20.65 [20.60, 20.70] & \textbf{20.44 [20.39, 20.49]} \\

\midrule
\multicolumn{5}{c}{\textbf{MARD} (\%)} \\
\midrule
GluForecast & 11.13 [11.10, 11.15] & 10.90 [10.88, 10.93] & \textbf{10.76 [10.74, 10.78]} & 10.76 [10.74, 10.79] \\
LightGBM    & 11.45 [11.42, 11.47] & 11.34 [11.32, 11.36] & 11.32 [11.30, 11.35] & \textbf{11.27 [11.25, 11.30]} \\
LSTM        & 11.08 [11.06, 11.11] & 11.46 [11.43, 11.48] & \textbf{10.82 [10.80, 10.84]} & 10.89 [10.87, 10.91] \\
Ridge       & 12.03 [12.00, 12.05] & 11.95 [11.93, 11.97] & 12.02 [12.00, 12.05] & \textbf{11.94 [11.91, 11.96]} \\
UniTS       & 11.09 [11.06, 11.11] & 10.99 [10.97, 11.02] & 11.05 [11.03, 11.07] & \textbf{10.93 [10.91, 10.95]} \\
\midrule
\multicolumn{5}{c}{\textbf{DTS A} (\%)} \\
\midrule

GluForecast & 86.75 [86.68, 86.81] & 87.31 [87.24, 87.36] & 87.58 [87.52, 87.65] & \textbf{87.65 [87.59, 87.71]} \\
LightGBM    & 86.14 [86.08, 86.21] & 86.45 [86.39, 86.52] & 86.40 [86.34, 86.47] & \textbf{86.55 [86.49, 86.61]} \\
LSTM        & 86.92 [86.86, 86.99] & 86.34 [86.28, 86.41] & \textbf{87.55 [87.49, 87.61]} & 87.44 [87.37, 87.50] \\
Ridge       & 85.12 [85.06, 85.19] & 85.33 [85.27, 85.39] & 85.13 [85.07, 85.20] & \textbf{85.34 [85.28, 85.41]} \\
UniTS       & 86.93 [86.87, 86.99] & 87.09 [87.03, 87.15] & 86.90 [86.84, 86.96] & \textbf{87.25 [87.19, 87.31]} \\
\bottomrule
\end{tabular}%
}
\end{table*}

\section{Discussion}

MetaboNet-Bench addresses gaps in the literature by providing an updated, end-to-end benchmark that enables reproducible and extensible evaluation and comparison across diverse glucose prediction models in type 1 diabetes. The platform facilitates testing of models across tasks, modalities, and evaluation criteria. Our baseline experiments span classical methods, time-series foundation models, and multimodal approaches (including a novel multimodal algorithm), and illustrate how standardized reporting can clarify model strengths and limitations.

The utility of MetaboNet-Bench is highlighted through the performance differences between  standard metrics (RMSE and MARD) and clinically relevant metrics. Of note, we show how there are discrepancies in performance that are amplified postprandially and after insulin corrections. Further, we highlight that the best-performing model changes depending on the glycemic range. This is important as it suggests traditional metrics mask clinically meaningful implications. MetaboNet-Bench is fully open-source and highly extensible upon download of openly-accessible MetaboNet data. We encourage the community to add their models and share their results using the same splits provided by MetaboNet-Bench.

 Through our evaluation of the implemented models, we identified areas for the research community to focus on. First, current algorithms do not perform evenly across all glycemic ranges. This is especially important due to safety implications of incorrect predictions at different reference ranges. Second, our results emphasize that existing models struggle to perform well at longer time horizons and after system disturbances. Future work should focus on models designed for these tasks, and models which are expressive enough to take advantage of non-CGM inputs.  
 
\subsection{Limitations}
MetaboNet-Bench has several limitations. Our benchmark does not test models that incorporate data other than CGM, carbohydrate, and insulin data. Behavioral, physiological (e.g., wearable-derived), and other data types are known to impact glucose levels over time, but they are frequently unavailable or recorded intermittently.  Future work should seek to utilize the exercise data available in MetaboNet to see if models can benefit from this signal as well. In addition, the data tested are a subset of available data in MetaboNet, which includes only data with CGM, carbohydrate, and insulin data. Thus, the data on which the model is evaluated might be a biased sample of a T1D population. It is also possible that there was data missingness not accounted for in the modeling.  Our ablation study is also limited by the inclusion of only 5 models and the omission of a version of GluFormer (the largest model of the group) which takes both insulin and carbohydrate data.   Further, while we did conduct hyperparameter tuning on the models, it is possible that the results do not reflect completely optimized results of these architectures. We currently evaluate only generalized models for population-wide glucose forecasting. These models showed only slightly better performance on the known patients task. In the future, it may be possible to fine-tune individual patient models and evaluate them within this benchmark. This will require further dataset filtering since only a subset of patients are included in both MetaboNet-Bench test and train splits.

\subsection{Broader Impacts}

While the intention of an open-source benchmark for evaluating glucose forecasting models is to promote transparency and reproducibility in the field, we recognize there may be unintended harms with improper use of this resource. Namely, forecasting algorithms in T1D have the potential to cause great harm if used in the wrong context or used when making decisions on insulin dosing. While our benchmark aims to capture overall performance differences between models, our benchmark is not designed to evaluate the safety and robustness of forecasting algorithms in a deployment setting. This should be used for research purposes only as a starting point to understand differences in model performance and promote reproducibility in this space.

\section{Acknowledgments}
E.H. is supported by T32HD040128 from the NICHD/NIH. The content is solely the responsibility of the authors and does not necessarily represent the official views of the National Institutes of Health.
%\bibliography{paperpile,references}
%\bibliographystyle{icml2024}

\newpage
\appendix
\onecolumn

\section{Appendix - Datasets}

\begin{table}[ht]
\small
\caption{Overview of all datasets included in the public release of the MetaboNet dataset \citep{wolff2026metabonet}, showing the number of subjects per dataset. The column ``\# Not MDI'' reports the number of subjects using continuous insulin pump therapy, i.e., not treated with Multiple Daily Injections (MDI). The MetaboNet consolidated dataset is distributed under a custom license that restricts use to non-commercial research and educational purposes. Component datasets retain their original licenses, with studies hosted by the Jaeb Center for Health Research governed by the terms of use (JAEB ToU) and the remaining datasets released under the Creative Commons Attribution 4.0 International license (CC BY 4.0).}
\centering
\scriptsize
\setlength{\tabcolsep}{4pt}
\renewcommand{\arraystretch}{1.1}
\begin{tabularx}{\linewidth}{%
  >{\raggedright\arraybackslash}X
  >{\raggedright\arraybackslash}X
  S[table-format=4.0]
  S[table-format=3.0]
  >{\centering\arraybackslash}c
  >{\raggedright\arraybackslash}l
}
\toprule
\textbf{Dataset Name} &
\textbf{Citation} &
\multicolumn{1}{c}{\textbf{\# Subjects}} &
\multicolumn{1}{c}{\textbf{\# Not MDI}} &
\textbf{Carbs Available} &
\textbf{License} \\
\midrule
CTR3 & \cite{Zisser2014ClosedLoopChallenges} & 30  & 30 & $\checkmark$ & JAEB ToU \\
DCLP3 & \cite{Brown2019SixMonthClosedLoop} & 112 & 112 & $\times$ & JAEB ToU \\
DCLP5 & \cite{Breton2020ClosedLoopChildren} & 100 & 100 & $\times$ & JAEB ToU \\
Flair & \cite{Bergenstal2021FLAIR} & 113 & 113 & $\times$ & JAEB ToU \\
IOBP2 & \cite{Lynch2022BionicPancreas} & 332 & 332 & $\times$ & JAEB ToU \\
Loop Observational Study & \cite{Lum2021LoopRealWorld} & 845 & 845 & $\checkmark$ & JAEB ToU \\
PEDAP & \cite{Wadwa2023PEDAP} & 65  & 65 & $\checkmark$ & JAEB ToU \\
ReplaceBG & \cite{Aleppo2017REPLACEBG} & 208 & 208 & $\checkmark$ & JAEB ToU \\
AZT1D & \cite{Khamesian2025-bh} & 23  & 23 & $\checkmark$ & CC BY 4.0 \\
BrisT1D & \cite{James2025BrisT1DOpenDataset} & 19 & 19 & $\checkmark$ & CC BY 4.0 \\
HUPA-UCM & \cite{HidalgoAlvaradoBotella2024HUPAUCM} & 22 & 12 & $\checkmark$ & CC BY 4.0 \\
Shanghai T1DM & \cite{zhao2022_shanghaiT1DM_T2DM} & 12 &  9    & $\checkmark$ & CC BY 4.0 \\
T1D-UOM & \cite{AlsuhaymiBilalGarcia2025LongitudinalMultimodalDataset} & 14 & 6 & $\checkmark$ & CC BY 4.0 \\
\hline
\textbf{Total} & & \textbf{1895} & \textbf{1874} & & \\
\bottomrule
\end{tabularx}
\label{tab:datasets}
\end{table}

\section{Appendix — Implementation Details}
\label{app:methods}
\subsection{Training details and hyperparameters}
\textbf{Gluformer:}
Gluformer was trained following instructions on the GitHub \citep{Sergazinov2022-oe}. Gluformer was trained using the Adam optimizer with a learning rate of 0.001 for 50 epochs with negative log-likelihood loss over a mixture distribution of future glucose trajectories. These parameters were selected based on recommended defaults in the paper and associated open source code. Training uses contiguous 16-hour sequences from the dataset, with 15 hours of input and 1 hour of target values. We used the log sum likelihood loss described in \cite{Sergazinov2024}. To match the paper, we generated 5-dimensional time embeddings for each input and predicted sample. The time embedding consists of day of year, day of month, day of week, hour of day and minute of hour, each normalized to [0, 1].

\textbf{LightGBM:}
For LightGBM, sequences from the MetaboNet training set were converted into a flat tabular representation with two hours of time-lagged features (history length $H = 24$ at the dataset's 5-minute sampling rate), with windows generated using non-overlapping episodes. The prediction horizons were fit jointly via scikit-learn's \texttt{MultiOutputRegressor}, which trains one LightGBM regressor per horizon, tuned with $100$ trials of Optuna using the default Tree-structured Parzen Estimator (TPE) sampler over a custom-defined search space. For each trial the sampler proposed a configuration consisting of the number of boosting rounds $n_{\text{estimators}} \in [50, 500]$, the maximum tree depth $\in [3, 12]$, the learning rate $\in [0.01, 0.3]$ on a logarithmic scale, the maximum number of leaves per tree $\in [10, 300]$, the per-tree feature subsampling fraction $\in [0.5, 1.0]$, the per-tree row subsampling (bagging) fraction $\in [0.5, 1.0]$, the bagging frequency $\in [0, 7]$ iterations, the minimum number of samples per leaf $\in [5, 100]$, and the
$L_1$ and $L_2$ leaf-weight regularization strengths $\lambda_1, \lambda_2 \in [0, 10]$. Each candidate was scored by the mean root
mean squared error (RMSE) under 3-fold cross-validation on the training split. The configuration achieving the lowest mean CV-RMSE was retained and refit on the full training split before evaluation. The random seed was fixed to $42$.

The Optuna-selected configurations differ across ablations. With the full feature set (CGM + Carbs + Insulin), the best configuration uses $474$ boosting rounds at depth $10$ with $277$ leaves, learning rate $0.079$, feature fraction $0.78$, bagging fraction $0.59$ at frequency $0$, $38$ samples per leaf, and regularization $\lambda_1 = 6.73$, $\lambda_2 = 7.83$. The
CGM\,+\,Carbs ablation uses $480$ rounds at depth $12$ with $294$ leaves, learning rate $0.060$, feature fraction $0.90$, bagging fraction $0.81$ at frequency $2$, $95$ samples per leaf, and $\lambda_1 = 6.40$, $\lambda_2 = 8.98$. The CGM-only ablation uses $388$ rounds at depth $12$ with $284$ leaves, learning rate $0.066$, feature fraction $0.86$, bagging fraction $0.98$ at frequency $3$, $64$ samples per leaf, and $\lambda_1 = 8.27$, $\lambda_2 = 2.45$. Finally, the CGM\,+\,Insulin ablation uses $454$ rounds at depth $11$ with $239$ leaves, learning rate $0.084$,
feature fraction $0.78$, bagging fraction $0.72$ at frequency $0$, $91$ samples per leaf, and $\lambda_1 = 0.22$, $\lambda_2 = 2.28$.

\textbf{UniTS:}
The UniTS model was prompt-tuned using the Adam optimizer with a batch size of 256 and learning rate and weight decay recommended by Optuna. We used 30 trials of Optuna using the default TPE sampler over learning rates $\in [0.00001, 0.001]$ and weight decays $\in [0.00001, 0.001]$. Each ablation was tuned separately. With the full feature set (CGM + Insulin + Carbs) the best configuration uses a learning rate of $0.00087$ and a weight decay of $0.0000102$. The CGM + Insulin ablation uses a learning rate of $0.00093$ and weight decay of $0.0000105$. The CGM + carbs ablation uses a learning rate of $0.00065$ and weight decay of $0.00033$. The CGM only ablation uses a learning rate of $0.00087$ and weight decay of $0.0000102$.

\textbf{LSTM:}
The LSTM models were trained using the Adam optimizer with MSE loss.  We use the tanh activation function for cell state updates and the sigmoid activation function for input, forget and output gates. Hyperparameters and their values were tuned identically to UniTS. With the full feature set (CGM + Insulin + Carbs) the best configuration uses a learning rate of $0.00098$ and a weight decay of $0.0000109$. The CGM + Insulin ablation uses a learning rate of $0.001$ and weight decay of $0.00001$. The CGM + carbs ablation uses a learning rate of $0.00096$ and weight decay of $0.000012$. The CGM only ablation uses a learning rate of $0.00087$ and weight decay of $0.0000102$.

\textbf{GluForecast:}
The GluForecast models were trained using the Adam optimizer with MSE loss. The model uses an inner dimension of $128$ with $4$ heads, $4$ transformer layers and a dropout of $0.1$ between each layer. Hyperparamters and their values were tuned identically to UniTS and LSTM. With the full feature set (CGM + Insulin + Carbs) the best configuration uses a learning rate of $0.00087$ and a weight decay of $0.0000108$. The CGM + Insulin ablation uses a learning rate of $0.00087$ and weight decay of $0.000088$. The CGM + carbs ablation uses a learning rate of $0.00092$ and weight decay of $0.000049$. The CGM only ablation uses a learning rate of $0.00048$ and weight decay of $0.000013$.

\textbf{Ridge:}
Ridge was trained with an identical preprocessing and hyperparameter tuning setup as LightGBM, with two differences: features were standardized to zero mean and unit variance via a scikit-learn \texttt{Pipeline} with \texttt{StandardScaler} before fitting, and the search space reduced to the single regularization strength $\alpha \in [0.1, 100]$ on a logarithmic scale. The Optuna-selected $\alpha$ was $1.378$ for the full feature set (CGM + Carbs + Insulin), $0.523$ for the CGM\,+\,Carbs ablation, $1.377$ for the CGM-only ablation, and $1.358$ for the CGM\,+\,Insulin ablation. 

\subsection{GluForecast}
\label{app:GluForecast}

We generated time embeddings with $4$ dimensions as follows:

Let $t$ denote a timestamp measured in seconds.
We define the day and week periods as:
\[
D = 24 \cdot 60 \cdot 60, \qquad W = 7D
\]
\[
\qquad {t}_{day} = t  \bmod D \qquad {t}_{week} = t \bmod W
\]

The angular phases are given by
\[
\theta_{\mathrm{day}}(t) = 2\pi \frac{{t}_{day}}{D},
\qquad
\theta_{\mathrm{week}}(t) = 2\pi \frac{{t}_{week}}{W}.
\]

The time embedding is then defined as
\[
\phi(t) =
\begin{bmatrix}
\sin\!\left(\theta_{\mathrm{day}}(t)\right) \\
\cos\!\left(\theta_{\mathrm{day}}(t)\right) \\
\sin\!\left(\theta_{\mathrm{week}}(t)\right) \\
\cos\!\left(\theta_{\mathrm{week}}(t)\right)
\end{bmatrix}
\]

This time embedding unambiguously expresses time in cyclical day and week patterns, giving the model the best chance to learn daily and weekly patterns affecting glucose.

During training, loss is calculated as follows:
\[
\mathcal{L}
= \sum_{h=1}^{H} w_h
\sum_{t=c}^{T-h}
\ell\!\left({y}_{t} + \hat{y}_{h,t},\, y_{t+h}\right),
\qquad
w_h = 10 + h .
\]

\noindent\textit{Horizon weight:}
We chose to weigh longer horizons more heavily to nudge the model to prioritize the more difficult prediction task. The constant $10$ balances the emphasis on long and short tasks. The moderately large horizon weights counteract the small deltas generated by the model.

\noindent\textit{Variables:}
$H$ is the number of prediction heads (horizons); $h \in \{1,\dots,H\}$ indexes the horizon (in 5 minute increments);
$t$ indexes time (in 5 minute increments); $c$ is the first time index included in the loss to allow a minimum context;
$T$ is the final time index of the sequence;
$y_{t}$ is the ground-truth glucose value at time $t$;
$\hat{y}_{h,t}$ is the model predicted change in glucose for horizon $h$ made at time $t$;
$\ell(\cdot,\cdot)$ is the per-timestep loss;
and $w_h$ is the horizon-dependent weight applied to head $h$.

\subsection{Model Input}

\begin{table}[htbp]
\centering

\footnotesize
\caption{Model inputs}
\begin{tabular}{lccccc}
\toprule
Model & Timestamps & CGM & Insulin & Carbs & Input length \\
\midrule
GluForecast & X & X & X & X & 167 \\ \hline
Gluformer & X & X &  &  & 180 \\ \hline
LE &  & X &  &  & 1 \\ \hline
LightGBM & X & X & X & X & 24 \\ \hline
LSTM &  & X & X & X & 180 \\ \hline
Ridge & X & X & X & X & 24 \\ \hline
UniTS &  & X & X & X & 180 \\ \hline
ZOH &  & X &  & & 1 \\ 
\bottomrule
\end{tabular}

\label{tab:data}
\end{table}
\subsection{Model Complexity}
Table \ref{tab:complexity} shows the complexity of each model. 
\begin{table}[htbp]
\centering
\begin{small}
\begin{tabular}{lccc}
\hline
\toprule
Model & Params & FLOPs & MACs \\
\midrule
Gluformer & 11.2M & 2.727G & 1.308G \\ \hline
UniTS & 2.69M & 188.93M & 94.2M \\ \hline
GluForecast & 819K & 264.78M & 131.74M \\ \hline
LSTM & 202K & 71.8M & 1.54K \\ \hline
Ridge & 1048 & - & - \\ \hline
LightGBM & nonparametric & - & - \\ \hline

\end{tabular}
\end{small}
\caption{This table summarizes the model sizes and their runtime complexity, measured in Floating Point Operations (FLOPS) and Multiply-Accumulate Operations (MACs).}
\label{tab:complexity}
\end{table}

\subsection{Implementation Details and Hardware} 

All models were trained on a single H200 GPU with 141GB of VRAM unless otherwise mentioned. Each model was trained until evaluation loss plateaued. UniTS took about two hours to train for $20$ epochs. LSTM took about one hour to train for $20$ epochs. GluForecast took about $4$ hours to train for $40$ epochs. Other models were less computation resource-intensive and were trained on a CPU node with AMD EPYC 7301 16-Core Processor with 503GB memory. Full end to end evaluation including benchmarking, metrics calculation and figure generation takes about one hour on H200. The evaluation did not focus on execution efficiency for model training and testing. Thus, detailed information regarding memory consumption and time cost was not collected for reporting purposes. 

\section{Appendix - Metrics}
\label{app:metrics}
\textbf{Conventional Metrics:}
We selected RMSE and MARD because of their popularity for forecasting models. 
We reported model size as it dictates which systems will be capable of running each model, as well as FLOPS to indicate the computation complexity of each model and multiply-accumulate operations (MACs) to indicate how much common optimization libraries and dedicated acceleration hardware might speed up inference.

RMSE and MARD are calculated as follows:

\[
RMSE = \sqrt{\frac{1}{n}\sum_{i=1}^{n}{({y}_{i} - \hat{y}_{i})}^2}
\]

\[
MARD = \frac{1}{n}\sum_{1}^{n}{|{y_{i} - \hat{y}_{i}}|}
\]

where $n$ is the number of samples, ${y}_{i}$ is the reference value at location $i$ and ${\hat{y}_{i}}$ is the prediction at location $i$.

\textbf{Clinical Metrics:} The DTS error grid indicates the clinical significance or seriousness of model error \citep{dts-error-grid}. It is an industry-standard visualization that maps each prediction-reference pair to a clinically-defined risk zone based on the real-world implications of the error. The zone counts for the DTS error grid indicate how many of the predictions fall within each clinical zone - with zone A representing clinically safe predictions and zone B through E indicating errors with increasingly serious clinical consequences. Compared with alternate error grids (e.g., Clarke), the DTS framework provides a modern, expert-defined, and treatment-relevant characterization of risk. In our results, we report the percentage of time spent in each error zone, grouping C, D, and E together for reporting simplicity.

We used the following definitions of \textit{postprandial} and \textit{correction bolus} in order to produce figures in figures \ref{fig:correction_meal_diff}, \ref{fig:correction}, \ref{fig:mealtime}, and \ref{fig:scaling_laws}. 

\textbf{Postprandial:} For a given sample, we look at the current reported carbohydrates as well as the reported carbs every 5 minutes up until 30 minutes prior to the sample. If any of these carbs are nonzero we say the sample is \textit{postprandial}.

\textbf{Correction Bolus:} If a sample has a CGM value > 250 mg/dL and an insulin value of > 2IU for any of the intervals until 30 minutes prior to the sample, we call that sample a \textit{correction bolus}. 

\newpage
\section{Appendix — Supplementary Tables}

\begin{table*}[ht]
\centering
\caption{Performance comparison of glucose prediction models across different prediction horizons. The best (lowest) values for each horizon are highlighted in bold.}
\label{tab:overall-metrics}
\resizebox{\textwidth}{!}{
\begin{tabular}{lcccccccccccc}
\toprule
\multicolumn{13}{c}{\textbf{RMSE (mg/dL)}} \\
\midrule
PH (min)    & 5 & 10 & 15 & 20 & 25 & 30 & 35 & 40 & 45 & 50 & 55 & 60 \\
\midrule
GluForecast & \textbf{3.57} & \textbf{6.85} & \textbf{10.39} & \textbf{13.83} & \textbf{17.03} & \textbf{19.95} & \textbf{22.65} & \textbf{25.13} & \textbf{27.40} & \textbf{29.50} & \textbf{31.41} & \textbf{33.17} \\
Gluformer   & 5.28 & 8.24 & 11.86 & 15.38 & 18.82 & 21.97 & 24.87 & 27.53 & 29.94 & 32.11 & 34.13 & 36.35 \\
LightGBM    & 3.69 & 7.18 & 10.89 & 14.47 & 17.79 & 20.84 & 23.64 & 26.21 & 28.56 & 30.72 & 32.69 & 34.48 \\
LE          & 4.68 & 9.50 & 14.69 & 19.99 & 25.31 & 30.60 & 35.84 & 41.02 & 46.14 & 51.17 & 56.10 & 60.92 \\
LSTM        & 4.27 & 7.01 & 10.49 & 13.89 & 17.11 & 20.06 & 22.78 & 25.29 & 27.57 & 29.69 & 31.64 & 33.41 \\
Ridge       & 3.85 & 7.56 & 11.46 & 15.24 & 18.77 & 22.03 & 25.03 & 27.77 & 30.27 & 32.54 & 34.60 & 36.46 \\
UniTS       & 4.01 & 7.05 & 10.60 & 14.10 & 17.42 & 20.44 & 23.26 & 25.79 & 28.18 & 30.33 & 32.37 & 34.14 \\
ZOH         & 5.71 & 10.62 & 15.11 & 19.19 & 22.90 & 26.28 & 29.37 & 32.22 & 34.84 & 37.25 & 39.47 & 41.52 \\
\midrule
\multicolumn{13}{c}{\textbf{MARD (\%)}} \\
\midrule
PH (min)    & 5 & 10 & 15 & 20 & 25 & 30 & 35 & 40 & 45 & 50 & 55 & 60 \\
\midrule
GluForecast & \textbf{1.69} & \textbf{3.47} & \textbf{5.42} & \textbf{7.27} & 9.11 & \textbf{10.76} & \textbf{12.33} & \textbf{13.84} & 15.23 & 16.57 & 17.78 & 18.91 \\
Gluformer   & 2.60 & 4.10 & 6.09 & 8.06 & 9.99 & 11.69 & 13.31 & 14.73 & 16.03 & 17.32 & 18.52 & 19.57 \\
LightGBM    & 1.76 & 3.67 & 5.71 & 7.68 & 9.54 & 11.27 & 12.90 & 14.43 & 15.85 & 17.18 & 18.41 & 19.55 \\
LE          & 2.32 & 4.83 & 7.47 & 10.15 & 12.84 & 15.54 & 18.25 & 20.94 & 23.62 & 26.27 & 28.90 & 31.50 \\
LSTM        & 2.31 & 3.72 & 5.57 & 7.45 & \textbf{9.11} & 10.89 & 12.40 & \textbf{13.81} & \textbf{15.06} & \textbf{16.39} & \textbf{17.54} & \textbf{18.58} \\
Ridge       & 1.79 & 3.81 & 5.97 & 8.06 & 10.05 & 11.94 & 13.71 & 15.37 & 16.91 & 18.33 & 19.65 & 20.86 \\
UniTS       & 1.97 & 3.62 & 5.56 & 7.43 & 9.34 & 10.93 & 12.49 & 13.91 & 15.37 & 16.49 & 17.73 & 18.79 \\
ZOH         & 2.91 & 5.49 & 7.85 & 10.00 & 11.98 & 13.80 & 15.49 & 17.06 & 18.51 & 19.86 & 21.12 & 22.29 \\
\bottomrule
\end{tabular}
}
\end{table*}

\begin{table*}[t]
\centering
\caption{Performance comparison on the \textbf{Task 1: Novel Patients split} of glucose prediction models across different horizons and metrics. The best (lowest) values for each horizon are highlighted in bold. This captures how well models generalize to new patients.}
\label{tab:person-independent-split}
\resizebox{\textwidth}{!}{
\begin{tabular}{lcccccccccccc}
\toprule
\multicolumn{13}{c}{\textbf{RMSE (mg/dL)}} \\
\midrule
PH (min)    & 5 & 10 & 15 & 20 & 25 & 30 & 35 & 40 & 45 & 50 & 55 & 60 \\
\midrule
GluForecast & \textbf{3.58} & \textbf{6.88} & \textbf{10.46} & \textbf{13.91} & \textbf{17.13} & \textbf{20.06} & \textbf{22.76} & \textbf{25.23} & \textbf{27.50} & \textbf{29.59} & \textbf{31.49} & \textbf{33.23} \\
Gluformer   & 5.30 & 8.24 & 11.91 & 15.45 & 18.91 & 22.05 & 24.96 & 27.60 & 30.00 & 32.16 & 34.17 & 36.39 \\
LightGBM    & 3.71 & 7.22 & 10.96 & 14.55 & 17.89 & 20.95 & 23.75 & 26.31 & 28.65 & 30.79 & 32.75 & 34.51 \\
LE          & 4.72 & 9.58 & 14.83 & 20.18 & 25.56 & 30.89 & 36.17 & 41.38 & 46.52 & 51.57 & 56.52 & 61.36 \\
LSTM        & 4.29 & 7.04 & 10.55 & 13.97 & 17.20 & 20.16 & 22.88 & 25.38 & 27.65 & 29.75 & 31.69 & 33.45 \\
Ridge       & 3.87 & 7.60 & 11.54 & 15.34 & 18.89 & 22.16 & 25.15 & 27.89 & 30.37 & 32.63 & 34.67 & 36.51 \\
UniTS       & 4.02 & 7.08 & 10.66 & 14.19 & 17.52 & 20.55 & 23.38 & 25.91 & 28.29 & 30.43 & 32.45 & 34.20 \\
ZOH         & 5.74 & 10.69 & 15.22 & 19.31 & 23.04 & 26.42 & 29.52 & 32.35 & 34.96 & 37.35 & 39.56 & 41.60 \\
\midrule
\multicolumn{13}{c}{\textbf{MARD (\%)}} \\
\midrule
PH (min)    & 5 & 10 & 15 & 20 & 25 & 30 & 35 & 40 & 45 & 50 & 55 & 60 \\
\midrule
GluForecast & \textbf{1.70} & \textbf{3.49} & \textbf{5.47} & \textbf{7.35} & 9.22 & \textbf{10.90} & \textbf{12.50} & 14.03 & 15.44 & 16.80 & 18.02 & 19.17 \\
Gluformer   & 2.62 & 4.13 & 6.15 & 8.15 & 10.11 & 11.83 & 13.48 & 14.92 & 16.24 & 17.55 & 18.75 & 19.82 \\
LightGBM    & 1.77 & 3.70 & 5.77 & 7.77 & 9.66 & 11.43 & 13.09 & 14.65 & 16.09 & 17.43 & 18.68 & 19.83 \\
LE          & 2.34 & 4.89 & 7.57 & 10.30 & 13.03 & 15.77 & 18.52 & 21.25 & 23.96 & 26.65 & 29.30 & 31.93 \\
LSTM        & 2.32 & 3.76 & 5.63 & 7.53 & \textbf{9.21} & 11.03 & 12.56 & \textbf{13.99} & \textbf{15.25} & \textbf{16.59} & \textbf{17.75} & \textbf{18.80} \\
Ridge       & 1.80 & 3.85 & 6.03 & 8.16 & 10.18 & 12.09 & 13.89 & 15.58 & 17.13 & 18.56 & 19.89 & 21.10 \\
UniTS       & 1.97 & 3.65 & 5.62 & 7.51 & 9.44 & 11.06 & 12.64 & 14.08 & 15.56 & 16.68 & 17.93 & 18.99 \\
ZOH         & 2.94 & 5.56 & 7.96 & 10.14 & 12.14 & 13.98 & 15.70 & 17.28 & 18.75 & 20.10 & 21.36 & 22.54 \\
\bottomrule
\end{tabular}
}
\end{table*}

\begin{table*}[t]
\centering
\caption{Performance comparison on the \textbf{Task 2: Known Patients split} dataset of glucose prediction models across different horizons and metrics. This captures how well the models perform when they are trained on earlier data from the same patient they are tested on. The best (lowest) values for each horizon are highlighted in bold.}
\label{tab:fine-tuning-split}
\resizebox{\textwidth}{!}{
\begin{tabular}{lcccccccccccc}
\toprule
\multicolumn{13}{c}{\textbf{RMSE (mg/dL)}} \\
\midrule
PH (min)    & 5 & 10 & 15 & 20 & 25 & 30 & 35 & 40 & 45 & 50 & 55 & 60 \\
\midrule
GluForecast & \textbf{3.52} & \textbf{6.77} & \textbf{10.24} & \textbf{13.64} & \textbf{16.79} & \textbf{19.70} & \textbf{22.38} & \textbf{24.87} & \textbf{27.16} & \textbf{29.29} & \textbf{31.24} & \textbf{33.03} \\
Gluformer   & 5.25 & 8.23 & 11.74 & 15.21 & 18.62 & 21.76 & 24.67 & 27.35 & 29.79 & 31.99 & 34.04 & 36.26 \\
LightGBM    & 3.65 & 7.10 & 10.74 & 14.28 & 17.56 & 20.59 & 23.39 & 25.98 & 28.34 & 30.54 & 32.56 & 34.39 \\
LE          & 4.59 & 9.31 & 14.36 & 19.54 & 24.73 & 29.92 & 35.08 & 40.18 & 45.23 & 50.22 & 55.11 & 59.89 \\
LSTM        & 4.24 & 6.93 & 10.36 & 13.72 & 16.90 & 19.82 & 22.55 & 25.07 & 27.37 & 29.54 & 31.53 & 33.33 \\
Ridge       & 3.79 & 7.46 & 11.28 & 15.01 & 18.50 & 21.74 & 24.72 & 27.48 & 30.01 & 32.34 & 34.45 & 36.36 \\
UniTS       & 3.97 & 6.99 & 10.46 & 13.91 & 17.18 & 20.17 & 22.98 & 25.52 & 27.93 & 30.12 & 32.17 & 34.00 \\
ZOH         & 5.62 & 10.46 & 14.87 & 18.89 & 22.57 & 25.94 & 29.03 & 31.90 & 34.54 & 37.00 & 39.26 & 41.34 \\
\midrule
\multicolumn{13}{c}{\textbf{MARD (\%)}} \\
\midrule
PH (min)    & 5 & 10 & 15 & 20 & 25 & 30 & 35 & 40 & 45 & 50 & 55 & 60 \\
\midrule
GluForecast & \textbf{1.67} & \textbf{3.40} & \textbf{5.29} & \textbf{7.08} & \textbf{8.85} & \textbf{10.44} & \textbf{11.93} & \textbf{13.38} & 14.72 & 16.04 & 17.21 & 18.31 \\
Gluformer   & 2.55 & 4.02 & 5.94 & 7.84 & 9.70 & 11.36 & 12.92 & 14.28 & 15.53 & 16.80 & 17.98 & 18.99 \\
LightGBM    & 1.73 & 3.59 & 5.56 & 7.46 & 9.25 & 10.91 & 12.47 & 13.94 & 15.30 & 16.60 & 17.80 & 18.91 \\
LE          & 2.27 & 4.69 & 7.24 & 9.82 & 12.41 & 15.01 & 17.62 & 20.22 & 22.81 & 25.41 & 27.97 & 30.49 \\
LSTM        & 2.28 & 3.65 & 5.44 & 7.25 & 8.86 & 10.57 & 12.03 & 13.39 & \textbf{14.61} & \textbf{15.91} & \textbf{17.05} & \textbf{18.08} \\
Ridge       & 1.76 & 3.73 & 5.81 & 7.84 & 9.76 & 11.58 & 13.29 & 14.89 & 16.39 & 17.80 & 19.09 & 20.29 \\
UniTS       & 1.95 & 3.56 & 5.44 & 7.25 & 9.10 & 10.63 & 12.14 & 13.52 & 14.93 & 16.04 & 17.27 & 18.32 \\
ZOH         & 2.83 & 5.32 & 7.60 & 9.68 & 11.60 & 13.37 & 15.01 & 16.54 & 17.96 & 19.30 & 20.54 & 21.70 \\
\bottomrule
\end{tabular}
}
\end{table*}

\clearpage

\section{Appendix — Supplementary Figures}

\subsection{Performance on Standard Metrics}

\begin{figure}[h]
    \centering
    \includegraphics[width=0.8\linewidth]{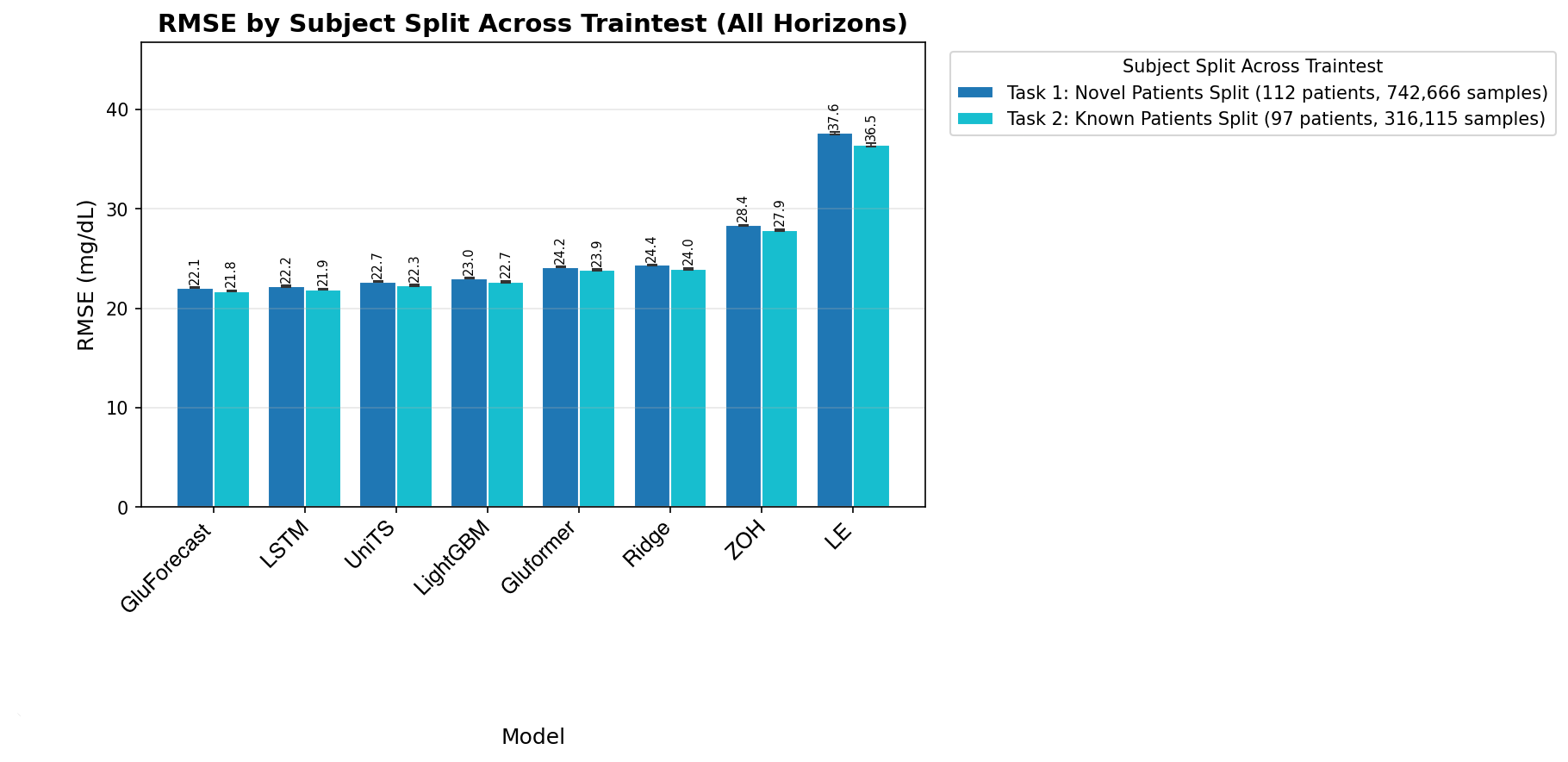}
    \caption{RMSE for Novel Patients, vs Known Patients with some data in the training set. The RMSE is calculated across all prediction horizons.  The MetaboNet dataset comes with a predefined train/test split.  In order to allow for testing generalization to new patients, within the test set, a portion of patients have their data split between train and test (known patients), and a portion do not (novel patients) \cite{wolff2026metabonet}. This figure shows that exposure to patients during training provides very minimal but consistent performance gains. Unless otherwise specified we draw from the \textit{full} test set in all other evaluations.}
    \label{fig:task-1-2}
\end{figure}

\begin{figure}[h]
    \includegraphics[width=0.8\linewidth]{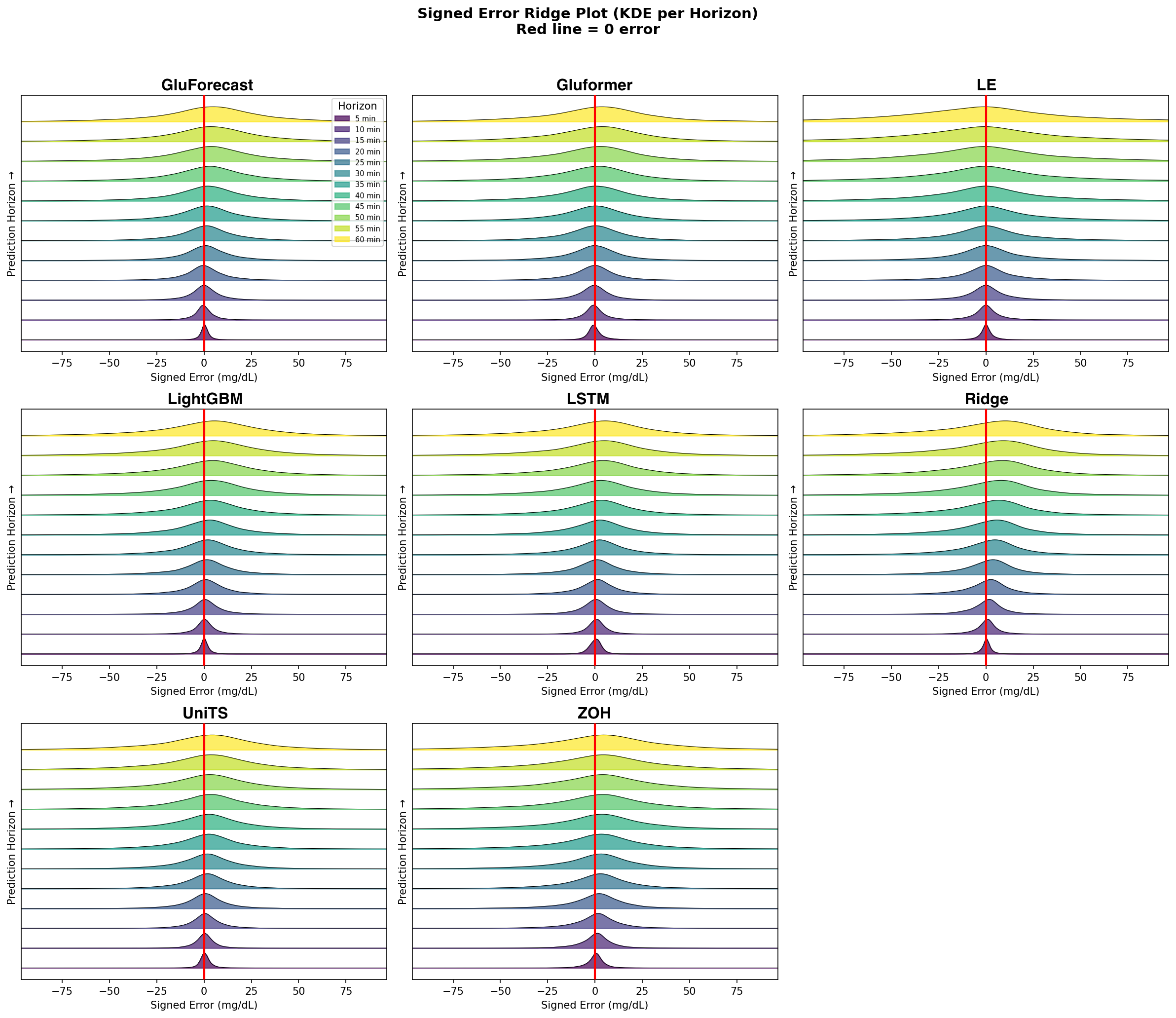}
    \caption{Illustration of the signed error for each model across prediction horizons, from shorter horizons at the bottom to longer horizons at the top. The Kernel Density Error (KDE) is visualized for each prediction horizon.}
    \label{fig:ridge-signed-error}
\end{figure}

\clearpage

\subsection{Evaluation Across Glycemic Profiles}

\begin{figure}[h]
    \centering
    \includegraphics[width=1\linewidth]{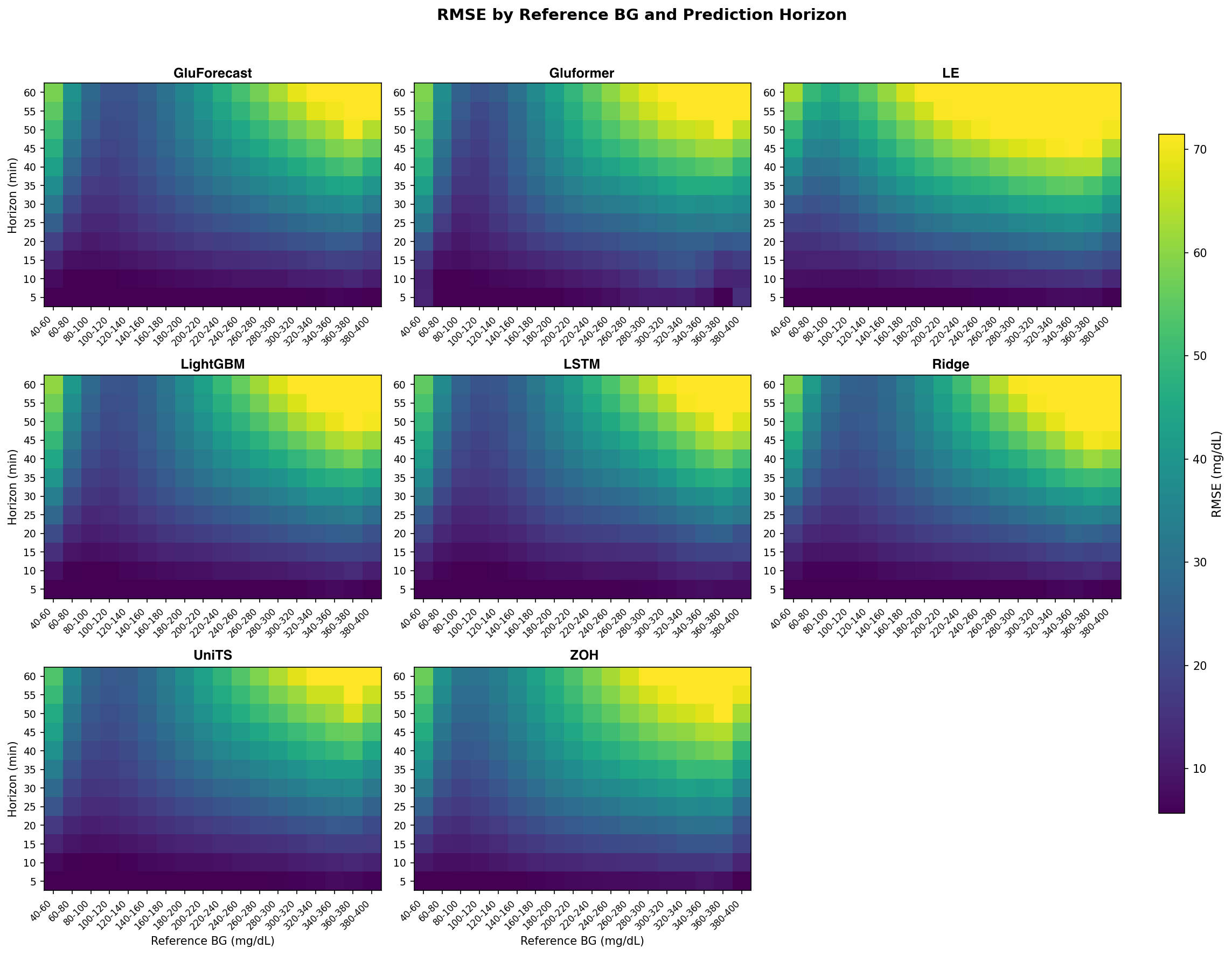}
\caption{Heatmap illustrating, for each model, the relationship between prediction horizon, reference blood glucose (BG) values, and RMSE. This illustrates which blood glucose ranges the models have trouble predicting and at which time horizons}
    \label{fig:heatmap-cgm-region}
\end{figure}

\clearpage
\subsection{Multimodal Performance at Mealtime and Correction}

\begin{figure}[h]
    \centering
    \includegraphics[width=0.92\linewidth]{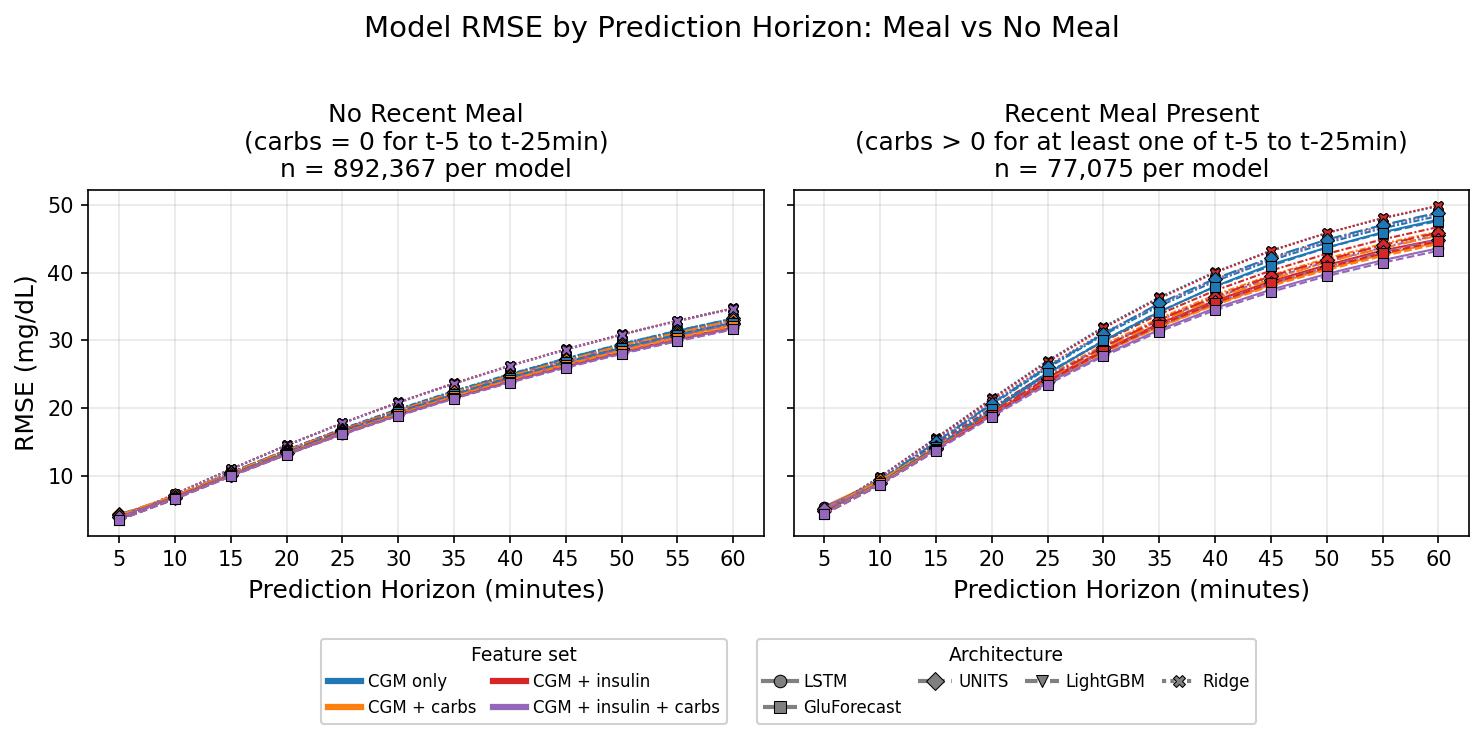}
    \caption{Comparison of model RMSE when there are no recent meals present in the data (left), compared to when there was a meal present during the previous 30 minutes (right).}
    \label{fig:mealtime}
\end{figure}

\begin{figure}[h]
    \centering
    \includegraphics[width=0.92\linewidth]{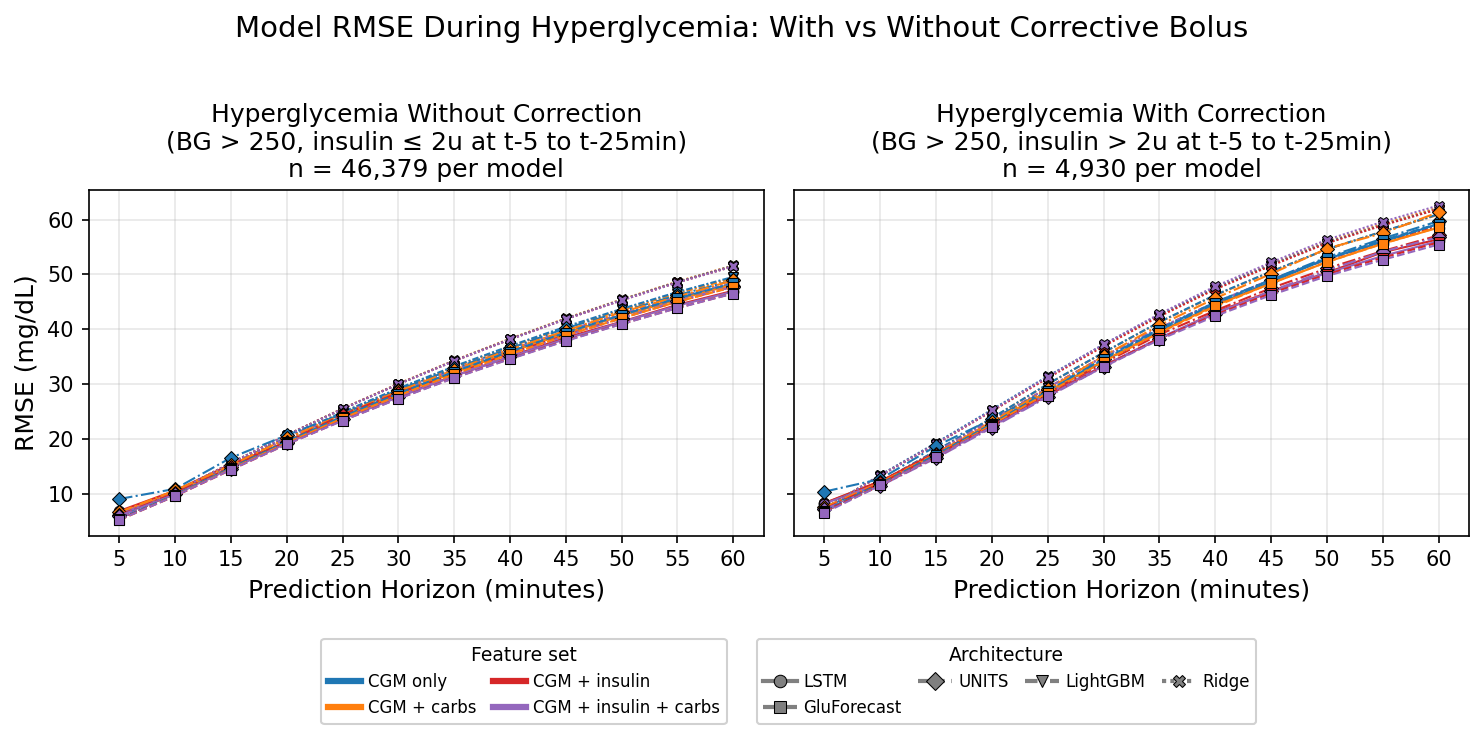}
    \caption{Comparison of model RMSE during hyperglycemic samples in the true values of the dataset. The left plot shows model performance when no correction insulin is given, while the right plot shows examples for when a dose larger than 2U of insulin is given within the previous 30 minutes.}
    \label{fig:correction}
\end{figure}

We performed some analysis comparing model performance during particularly clinically difficult situations, namely, meals, and hyperglycemic correction insulin. This analysis shows that the multimodal GluForecast model performs notably better for post-prandial predictions (see Figure \ref{fig:mealtime}). The multimodal GluForecast also performs better than otherwise following corrections (Figure \ref{fig:correction}).


\clearpage

\subsection{Evaluation Across Subpopulations}

\begin{figure}[h]
    \centering
    \begin{subfigure}[b]{0.48\linewidth}
        \includegraphics[width=\linewidth]{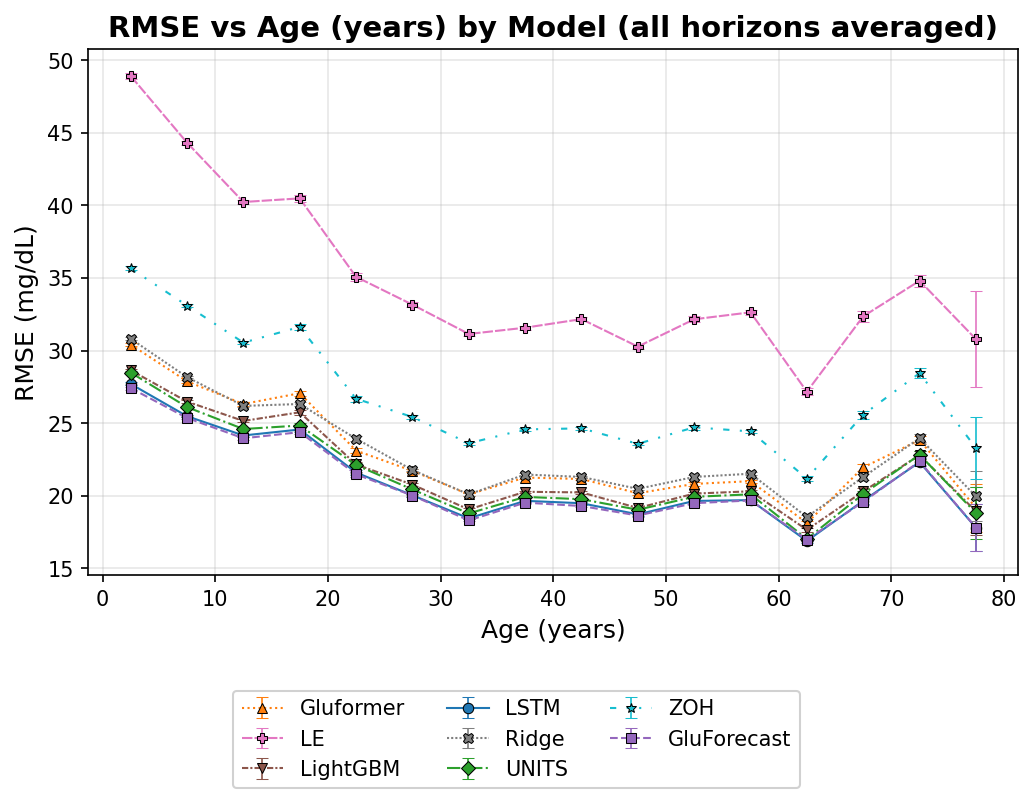}
        \caption{Age}
        \label{fig:subpop_age}
    \end{subfigure}
    \hfill
    \begin{subfigure}[b]{0.48\linewidth}
        \includegraphics[width=\linewidth]{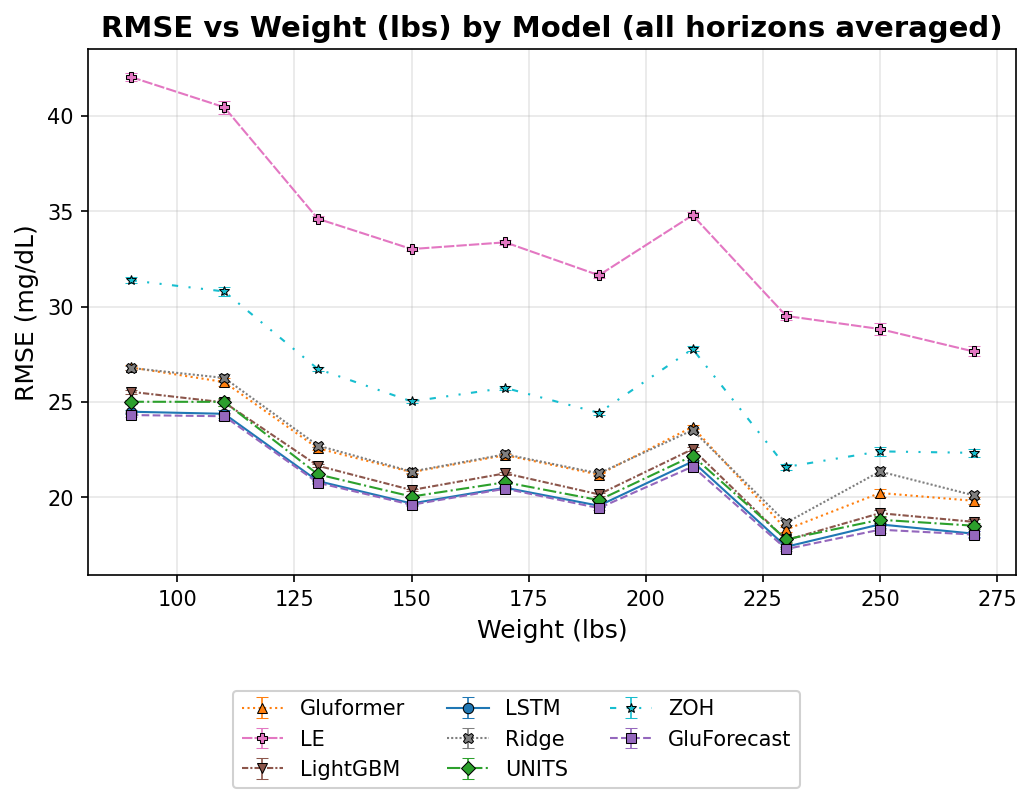}
        \caption{Weight}
        \label{fig:subpop_weight}
    \end{subfigure}
    
    \vspace{0.5em}
    
    \begin{subfigure}[b]{0.48\linewidth}
        \includegraphics[width=\linewidth]{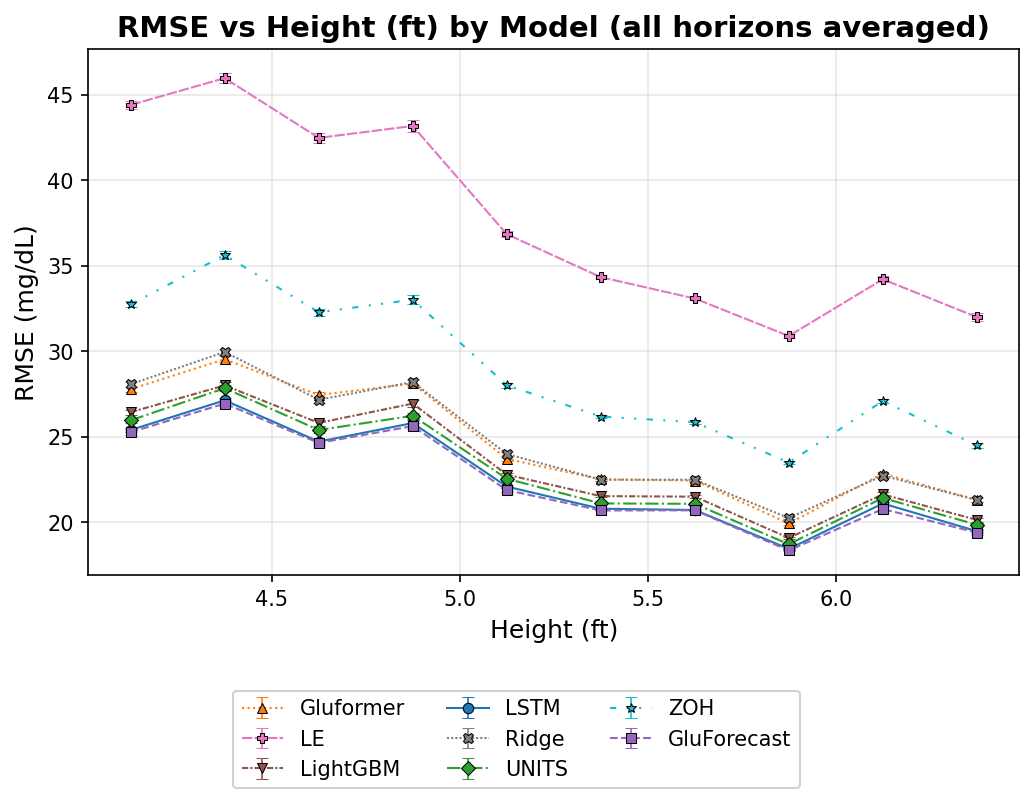}
        \caption{Height}
        \label{fig:subpop_height}
    \end{subfigure}
    \caption{RMSE results across subpopulations aggregated over all prediction horizons, illustrating the impact of age, weight, and height on model performance.}
    \label{fig:subpopulations_demographics}
\end{figure}

\begin{figure}[h]
    \centering
    \includegraphics[width=0.7\linewidth]{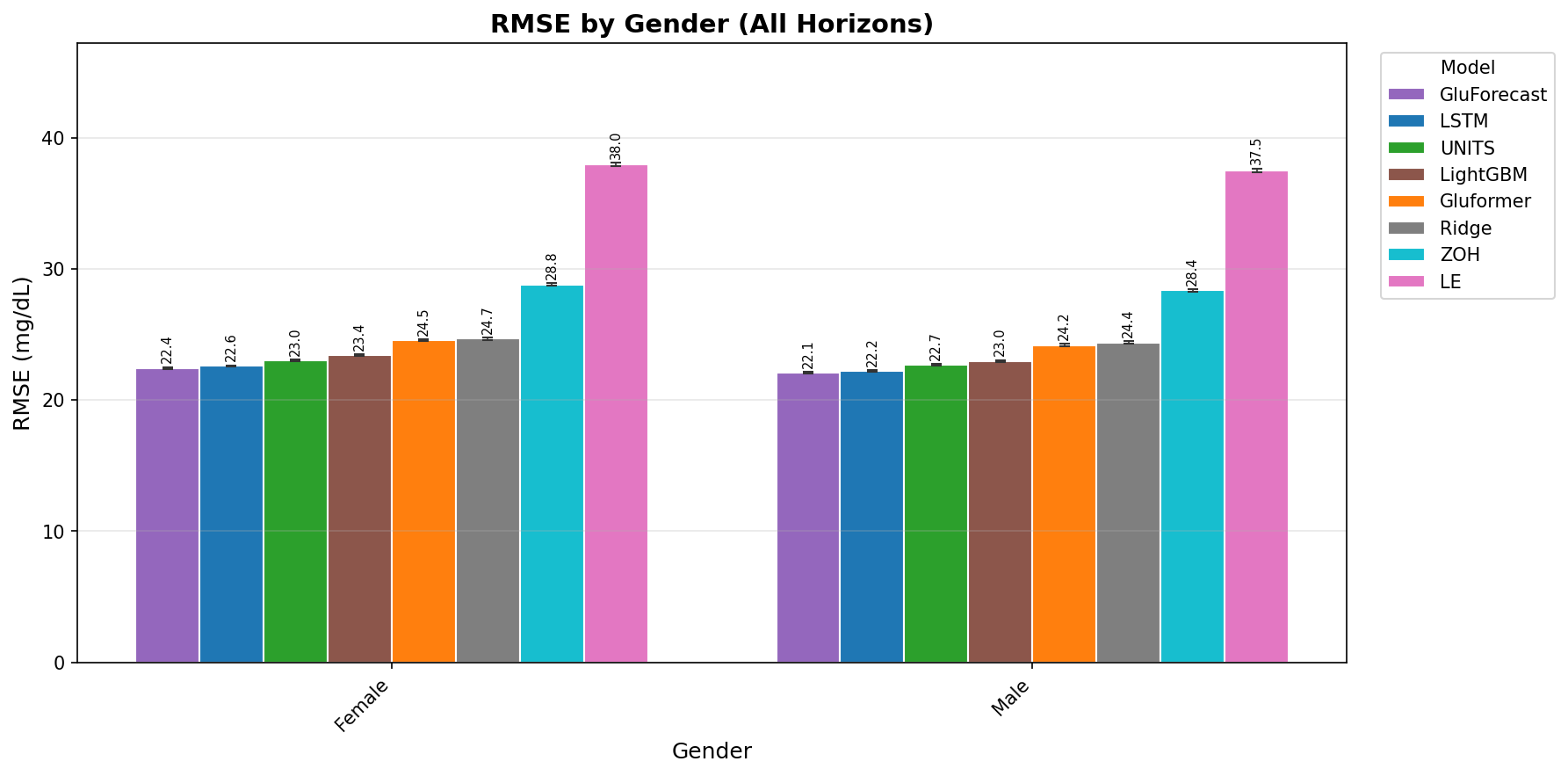}
    \caption{RMSE results across subpopulations aggregated over all prediction horizons, illustrating the impact of gender on model performance.}
    \label{fig:subpopulations_gender}
\end{figure}

\newpage

\end{document}